\pdfoutput=1

\documentclass[11pt]{article}

\usepackage[final]{acl}

\author{\textbf{Zhanchao Zhou}$^{1,2,3}$~~
    \textbf{Tianyi Wu}$^{4{\dag}}$\thanks{Equal Contribution; \dag Work done during internship at Westlake University; $\diamond$ Corresponding author.}~~
    \textbf{Zhiyun Jiang}$^{5{\dag}*}$~~
    \textbf{Fares Obeid$^{6}$~~}
    \textbf{Zhenzhong Lan}$^{2{\diamond}}$\\
    $^{1}$Zhejiang University~~~~~
    $^{2}$Westlake University~~~~~
    $^{3}$Ant Group\\
    $^{4}$University of Electronic Science and Technology of China \\
    $^{5}$China University of Mining and Technology ~~~ $^{6}$Imperial College London \\
}

\usepackage{times}
\usepackage{latexsym}

\usepackage[T1]{fontenc}

\usepackage[utf8]{inputenc}

\usepackage{microtype}

\usepackage{inconsolata}

\usepackage{graphicx}

\usepackage{amsmath}
\usepackage{amssymb}
\usepackage{subcaption}
\usepackage{bm}
\usepackage{colortbl}
\definecolor{gray}{RGB}{200,200,200}
\definecolor{lightyellow}{RGB}{255,255,224}
\definecolor{lightgray}{RGB}{234,234,234}
\usepackage{makecell}
\usepackage{tabularx}
\usepackage{booktabs}
\usepackage{arydshln}
\usepackage{colortbl}
\newcommand{\hdashlineself}{\arrayrulecolor{white}\hline\hline\arrayrulecolor{black}}
\usepackage{multirow}

\title{Value Residual Learning}

\begin{document}
\maketitle
\begin{abstract}
While Transformer models have achieved remarkable success in various domains, the effectiveness of information propagation through deep networks remains a critical challenge. Standard hidden state residuals often fail to adequately preserve initial token-level information in deeper layers. This paper introduces ResFormer, a novel architecture that enhances information flow by incorporating value residual connections in addition to hidden state residuals. And a variant is SVFormer, where all layers share the first layer's value embedding. Comprehensive empirical evidence demonstrates ResFormer achieves equivalent validation loss with 16.11\% fewer model parameters and 20.3\% less training data compared to Transformer, while maintaining similar memory usage and computational cost. Besides, SVFormer reduces KV cache size by nearly half with only a small performance penalty and can be integrated with other KV-efficient methods, yielding further reductions in KV cache, with performance influenced by sequence length and cumulative learning rate.
\end{abstract}

\section{Introduction}
The Transformer \citep{vaswani2017attention} model has become one of the leading architectures in recent years, excelling in both language modeling \citep{devlin2018bert, lan2019albert, brown2020language} and computer vision tasks \citep{dosovitskiy2020image}. Among its variants, decoder-only architectures have become the most prominent \citep{kaplan2020scaling, dubey2024llama}. The discovery of scaling laws \citep{hoffmann2022training, kaplan2020scaling} has driven the pursuit of larger Transformer models by increasing network depth and width. 

In a standard decoder-only transformer, initial token embeddings contain localized information, which rapidly evolves into abstract semantic features through early attention layers~\citep{Sun2024painters, Clark2024bert}. As Transformers deepen, a critical question arises: \textbf{How effectively is the initial information propagated to deeper layers}? One common answer is that residual connections of hidden states ensure access to initial information throughout the network. However, some studies~\citep{zhou2021deepvit,shi2022revisiting} have identified that the smoothing effect of attention mechanisms leads to over-smoothing, where token representations become increasingly similar as the network deepens. This indicates that in deeper layers, sequence-level features become dominant, while token-level features are diluted. DenseFormer~\citep{pagliardini2024denseformer} applied the idea of learnable dense connections from DenseNet~\citep{huang2016deep} to Transformer, and the learned connection coefficients shows that deeper layers indeed require larger attention to initial embeddings. Given the low similarity between initial token embeddings and deeper hidden states~\citep{Sun2024painters}, their directly summation may significantly impact the modeling of attention distribution for abstract semantic information in later layers. NeuTRENO~\citep{nguyen2023mitigating} alleviates over-smoothing from the view of regularizers by considering the difference between value vectors of the first and current layers.

\begin{figure*}[!h]
  \centering
\begin{subfigure}{0.19\linewidth}
\centering
\includegraphics[trim=0 0 5 0,clip,width=\linewidth]{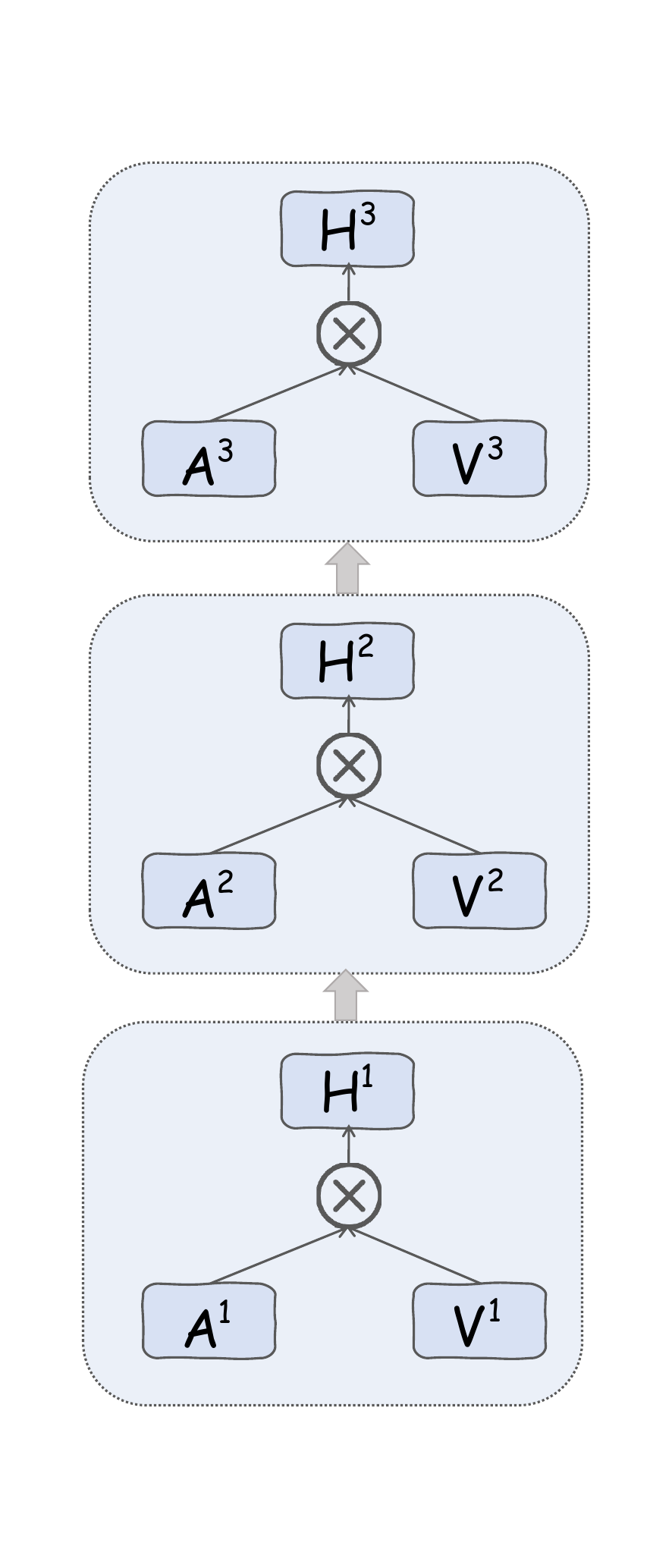}
\caption{Transformer}
\end{subfigure}
\begin{subfigure}{0.19\linewidth}
\centering
\includegraphics[trim=0 0 0 0,clip,width=\linewidth]{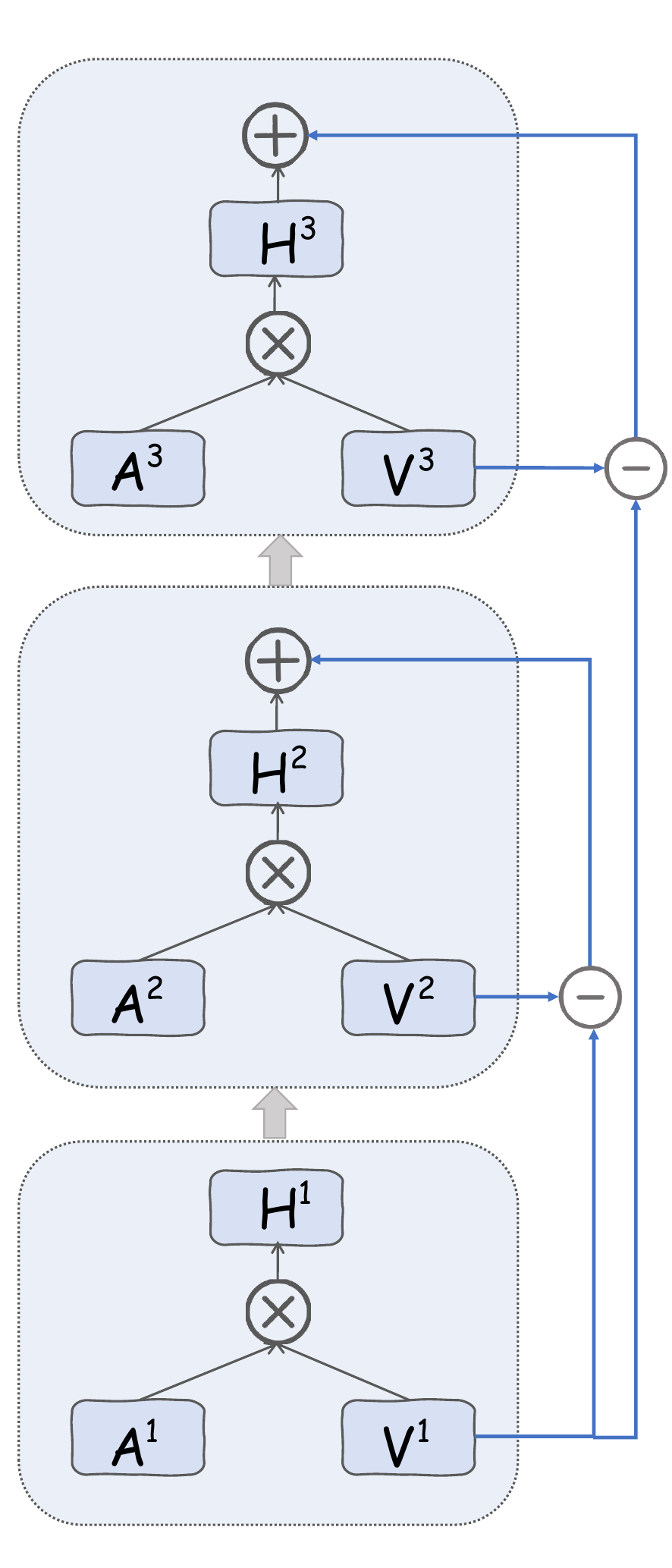}
\caption{NeuTRENO}
\end{subfigure}
\begin{subfigure}{0.19\linewidth}
\centering
\includegraphics[trim=0 0 0 0,clip,width=\linewidth]{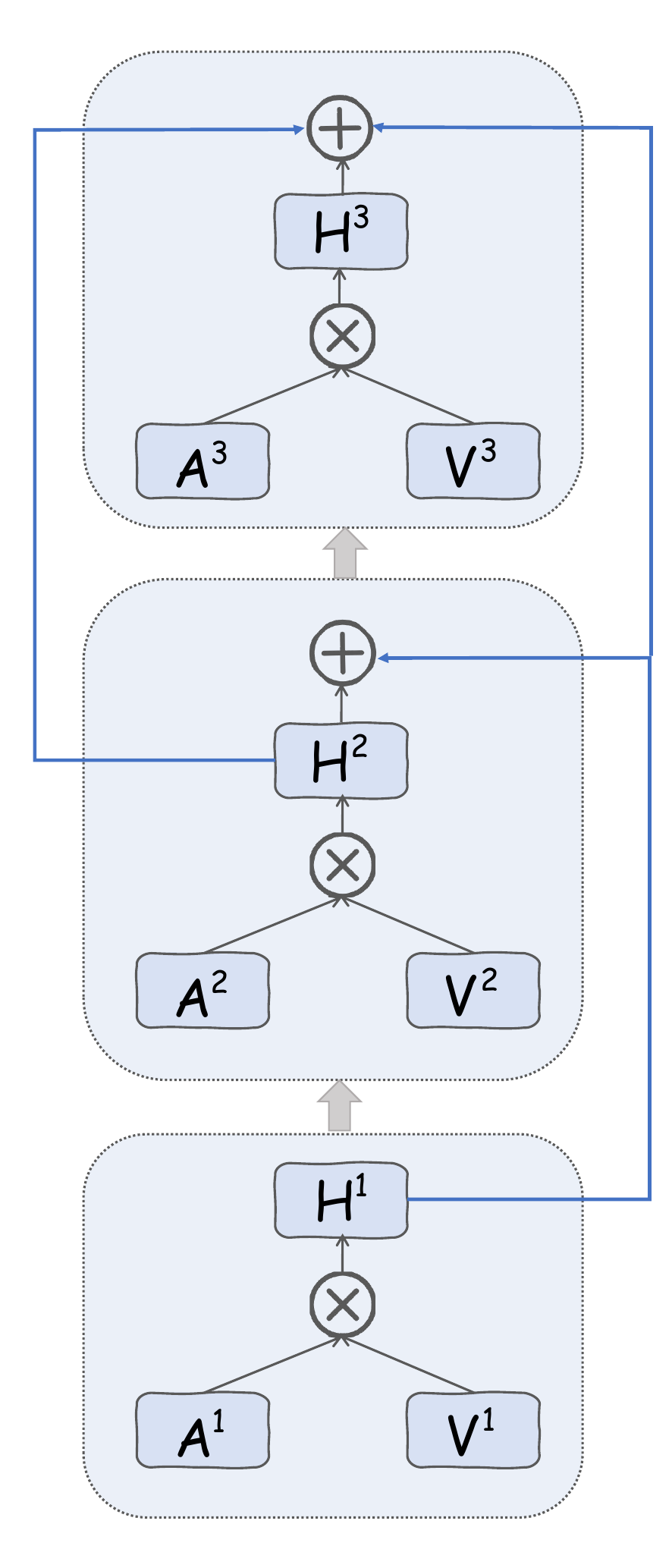}
\caption{DenseFormer}
\end{subfigure}
\begin{subfigure}{0.19\linewidth}
\centering
\includegraphics[trim=0 0 0 0,clip,width=\linewidth]{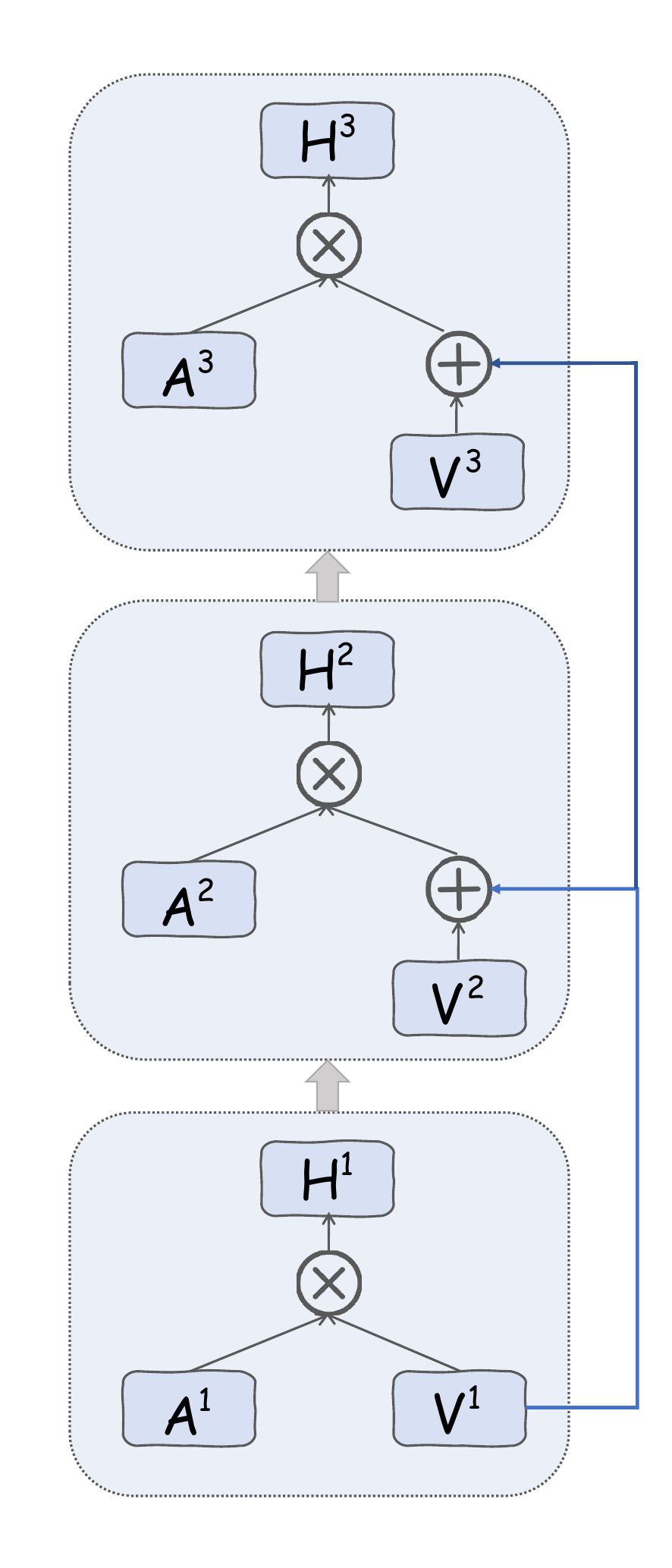}
\caption{ResFormer}
\end{subfigure}
\begin{subfigure}{0.19\linewidth}
\centering
\includegraphics[trim=0 0 0 0,clip,width=\linewidth]{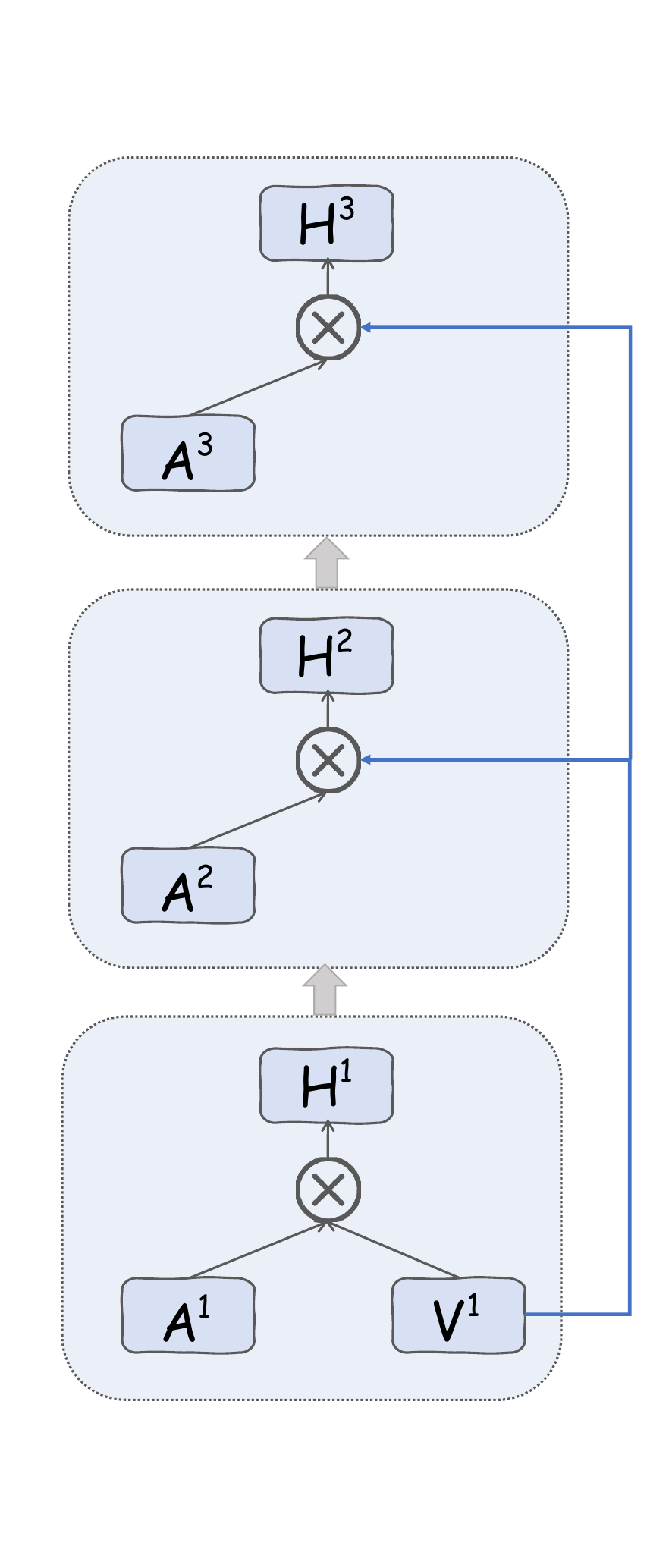}
\caption{SVFormer}
\end{subfigure}
\caption{Simplified illustration of the vanilla Transformer, NeuTRENO, DenseFormer, ResFormer, and SVFormer, with only three-layer structures and no operations other than attention. $\mathbf{A}^{i}$, $\mathbf{V}^{i}$, and $\mathbf{H}^{i}$ denote the attention matrix, value vectors, and attention outputs at the $i$-th layer, respectively. $\oplus$, $\ominus$, and $\otimes$ represent standard matrix addition, subtraction, and multiplication, respectively. 
}
\label{fig:Transformer_NeuTRENO_denseformer_ResFormer}
\end{figure*}

In this paper, we propose ResFormer, which enhances the propagation of initial local information by introducing value residual connections in addition to the standard hidden residual connections. Specifically, ResFormer applies a residual connection between the value vectors of the current layer and the first layer before the attention operation. In other words, both value states share the existing attention matrix of the current layer. The value states of the first attention layer and the preceding hidden states differ only by a linear transformation along the channel dimension, both representing token-level raw information. We hypothesize that introducing residual connections for values has a less impact on modeling attention distributions for sequence-level semantic information in higher layers and complements the original hidden state residual. Fig.~\ref{fig:Transformer_NeuTRENO_denseformer_ResFormer} illustrates a comparison of the extra skip connections introduced by different models.

During inference, deep networks require substantial $KV$ cache, severely impacting model deployment \citep{xiao2023efficient}. Existing $KV$-efficient methods often process keys and values simultaneously. Building on ResFormer, we decouple the value from the attention operation and propose a new kind of Transformer (SVFormer) where all layers share a single value state.

We experiment on a 20B SlimPajama sub-sampled dataset, using settings similar to popular large language models \citep{wei2023skywork, dubey2024llama, kaplan2020scaling}. We compare different models based on the valid loss against the vanilla Transformer. Results show that ResFormer outperforms the vanilla Transformer, DenseFormer, and NeuTRENO. ResFormer achieves equivalent validation loss with 16.11\% fewer model parameters and 20.3\% less training data compared to Transformer, while maintaining similar memory usage and computational cost. Besides, SVFormer, while reducing the $KV$-cache by nearly half, requires a 12.2\% increase in parameters to achieve the same validation loss as Transformer.
And SVFormer performs better when the training sequence length is longer. It further reduces the $KV$ cache when integrated with GQA~\citep{GQA}.

\section{Related Work}
\subsection{Shortcut Connections}
Deep learning models often consist of multiple layers, posing a challenge to minimize information loss during transmission. ResNet \citep{he2016deep} mitigates the vanishing gradient problem with identity connections. Stochastic Depth \citep{huang2016deep} enhances training by randomly dropping layers. DenseNet \citep{huang2017densely} allows subsequent layers to directly access the hidden states of all preceding layers. These two methods further enhance the information flow after ResNet. 

Related research indicates that, although increasing depth continues to yield performance improvements in language modeling tasks, the gains become less significant with further increases \citep{petty2024impact}. Furthermore, \cite{zhou2021deepvit} illustrates that a 32-layer ViT underperforms a 24-layer ViT. DenseFormer \citep{pagliardini2024denseformer} integrates weighted fusion of outputs from all preceding layers after each layer. To explore why increasing depth in Transformers does not yield expected gains, \cite{wang2022anti} finds that self-attention acts as a low-pass filter, smoothing token representations in ViTs. Additionally, \cite{shi2022revisiting} investigates over-smoothing from a graph perspective in BERT-based language modeling tasks. NeuTRENO \citep{nguyen2023mitigating} adds the difference between the value vectors of the first and current layers to each layer's attention output and significantly alleviates the over-smoothing problem. 

\subsection{\texorpdfstring{$KV$}{KV} cache compressing}
The KV cache significantly impacts the efficiency of long-text model inference, attracting extensive research. One category of Transformer-based methods addresses this by employing parameter or activation value sharing techniques. The most representative works include Multi-Query Attention \citep{MQA} and Grouped-Query Attention \citep{GQA} which suggest to share key and value across a group of queries. Besides, CLA \citep{brandon2024reducing} and LISA \citep{mu2024cross} respectively point out that we can reuse keys, values, or the attention matrix across layers to reduce redundancy between layers. While these methods typically process both key and value simultaneously, SVFormer is the first approach to decouple value from query and key during attention.

\section{Method}
\subsection{Preliminary}
\paragraph{Notations}
Let $\mathbf{H}_{n} \in \mathbb{R}^{l \times d}$ be the output hidden state of the $n$-th layer, where $l$ denotes the sequence length and $d$ is the dimension size. For each layer, the hidden state $\mathbf{H}_{n-1}$ will be firstly projected into $\mathbf{Q}_{n}, \mathbf{K}_{n}, \mathbf{V}_{n} \in \mathbb{R}^{l \times d}$ through three linear projections $\mathbf{W^{\mathbf{Q}}_n}, \mathbf{W^{\mathbf{K}}_n}, \mathbf{W^{\mathbf{V}}_n} \in \mathbb{R}^{d \times d}$ respectively. After these projections, the attention operation ($\operatorname{Attn}$), output projection ($\mathbf{W}^{\mathbf{O}}_n \in \mathbb{R}^{d \times d}$), and Multi-Layer-Perceptron ($\operatorname{Mlp}$) are applied sequentially: 
\begin{align}
\label{eqn:attention}
\mathbf{U}_{n}
&=\operatorname{Attn}(\mathbf{Q}_{n},\mathbf{K}_{n},\mathbf{V}_{n}). \\[0.1em]
\label{eqn:mlp}
\mathbf{H}_{n}
&=\operatorname{Mlp}(\mathbf{U}_n\mathbf{W}^{\mathbf{O}}_n).
\end{align}
\paragraph{NeuTRENO and DenseFormer}
After Eqn.~\ref{eqn:attention}, NeuTRENO adds the difference between the first and current layer's value:
\begin{align}
\label{eqn:NeuTRENO}
\mathbf{U}_{n}
&=\operatorname{Attn}(\mathbf{Q}_{n},\mathbf{K}_{n},\!\mathbf{V}_{n})+{\bm{\lambda}_n} (\mathbf{V}_{1}-\mathbf{V}_{n}).
\end{align}
After Eqn.~\ref{eqn:mlp}, DenseFormer performs a weighted average between all previous hidden states:
\begin{align}
\label{eqn:DenseFormer}
\mathbf{H}_{n}
&=\bm{\lambda}_{n,n}\operatorname{Mlp}(\mathbf{U}_n\mathbf{W}^{\mathbf{O}}_n)+\sum_{i = 0}^{n-1}\bm{\lambda}_{n,i}\mathbf{H}_{i}.
\end{align}
where $\mathbf{H}_{0}=\operatorname{Embedding}(\mathbf{X})$ for the input $\mathbf{X}$. $\bm{\lambda}_n$ in Eqn.~\ref{eqn:NeuTRENO} and $\!\{\bm{\lambda}_{n,i}\!\}_{i=0}^{n-1}$ in Eqn.~\ref{eqn:DenseFormer} are new parameters. Unless noted, $\bm{\lambda}_n$ is set to 0.4 for NeuTRENO suggested by \cite{nguyen2023mitigating}. For DenseFormer, only $\bm{\lambda}_{n,n}$ are set to 1 and the others are set to zero during initialization here.
\subsection{ResFormer}
\label{sec:resformer_formulation}
In contrast, before Eqn.~\ref{eqn:attention}, ResFormer introduces a skip connection from the first layer's value $\mathbf{V}_{1}=\mathbf{H}_{0}\mathbf{W}^{\mathbf{V}}_1$ to current layer’s value $\mathbf{V}_{n}=\mathbf{H}_{n-1}\mathbf{W}^{\mathbf{V}}_n$:
\begin{align}
\label{eqn:resformer}
\mathbf{V}_{n}
&={\bm{\lambda}_{n,1}}\mathbf{V}_{1}+{\bm{\lambda}_{n,2}}\mathbf{H}_{n-1}\mathbf{W}^{\mathbf{V}}_n.
\end{align}
where ${\bm{\lambda}_{n,1}}$ and ${\bm{\lambda}_{n,2}}$ are flexible scalars.

When all $\bm{\lambda}_{n,1}$ and $\bm{\lambda}_{n,2}$ are predetermined constants, it is termed \textbf{Constant-ResFormer}. If $\{{\bm{\lambda}_{n,1}} = {\bm{\lambda}_{n,2}}\}_{n=1}^N$, where $N$ is the total number of layers, the model is called \textbf{Identity-ResFormer}. Another variant where some layers have $\bm{\lambda}_{n,1} = 0$ is referred to as \textbf{Sparse-ResFormer}. Besides, if $\bm{\lambda}_{n,1}$ and $\bm{\lambda}_{n,2}$ are trainable parameters, the model is termed \textbf{Learnable-ResFormer}. Unless otherwise specified, $\bm{\lambda}_{n,1}$ and $\bm{\lambda}_{n,2}$ are initialized to 0.5 for Learnable-ResFormer and are predetermined as 0.5 for Identity-ResFormer. Furthermore, given that higher layers require more supplementary information from $\mathbf{V}_{1}$, we propose the \textbf{Learnable ResFormer Plus}. The initialization strategy is as follows: 1). $\bm{\lambda}_{n,2}$ is initialized to 0.5, where $n = 1, 2, \ldots, N$. 2). $\bm{\lambda}_{n,1}$ is initialized to $\bm{\lambda_{\text{scale}}} \cdot \frac{e^{\bm\lambda_{n,1}^{\prime}}}{\sum_{j=1}^{N} e^{\bm\lambda_{j,1}^{\prime}}}$, where $\bm{\lambda_{\text{scale}}}$ is initialized to the total layer number $N$ and is shared across all layers. 

A more general form is \textbf{Dense-ResFormer}, defined as $\mathbf{V}_{n}\!=\!\bm{\lambda}_{n,n}\mathbf{H}_{n-1}\mathbf{W}^{\mathbf{V}}_n+\sum_{i = 1}^{n-1}\bm{\lambda}_{n,i}\mathbf{V}_{i}$ for $n\geq2$, where $\!\{\bm{\lambda}_{n,i}\!\}_{i=1}^{n-1}$ are constants or trainable scalars. Unless noted, all $\bm{\lambda}_{n,n}$ are set to 1.
\subsection{SVFormer}
\begin{table}[!h]
\small
\centering
\begin{tabular}{ccccc}
\toprule
Shared Parts & - & values & keys & keys \& values \\
\midrule
Valid Loss & 2.739 & 2.743 & 2.753 & 2.776 \\
\bottomrule
\end{tabular}
\caption{Results of sharing different parts every 2 layers.}
\label{tab:CLA_ablation_kv}
\end{table}
\begin{figure*}[!htbp]
    \centering
      \begin{subfigure}{0.32\linewidth}
        \centering
        \includegraphics[width=\linewidth]{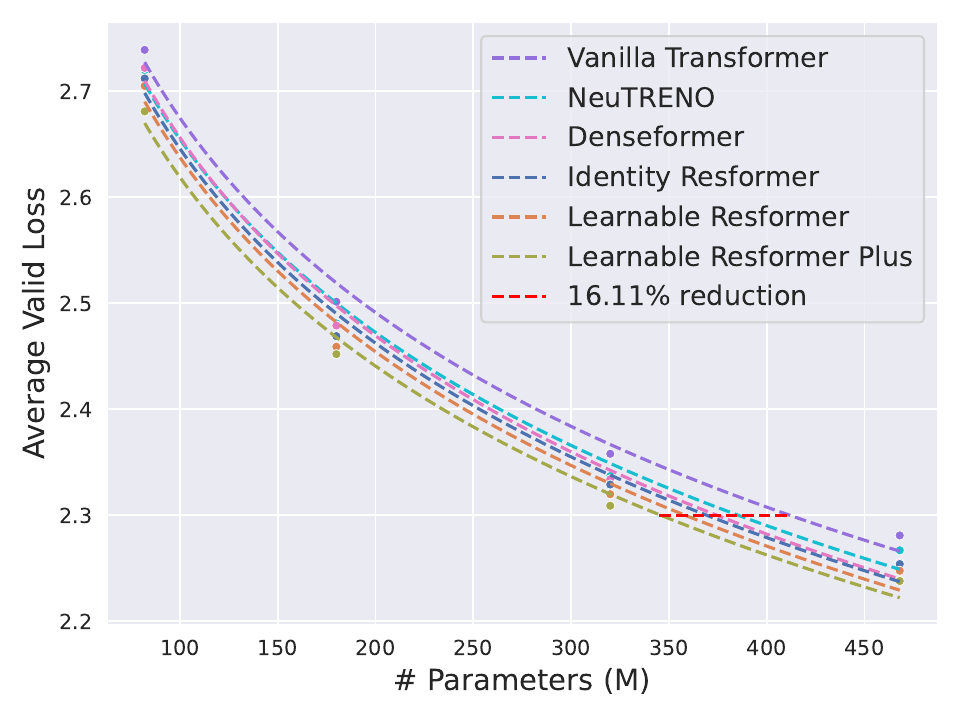}
      \end{subfigure}
      \begin{subfigure}{0.32\linewidth}
        \centering
        \includegraphics[width=\linewidth]{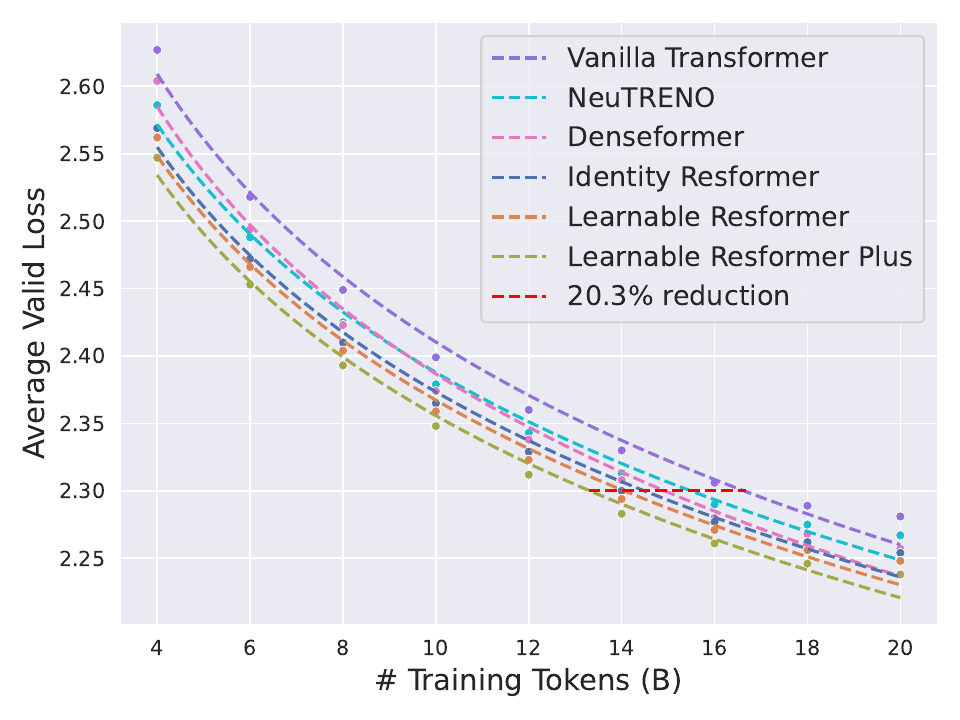}
      \end{subfigure}
      \begin{subfigure}{0.32\linewidth}
        \centering
        \includegraphics[width=\linewidth]{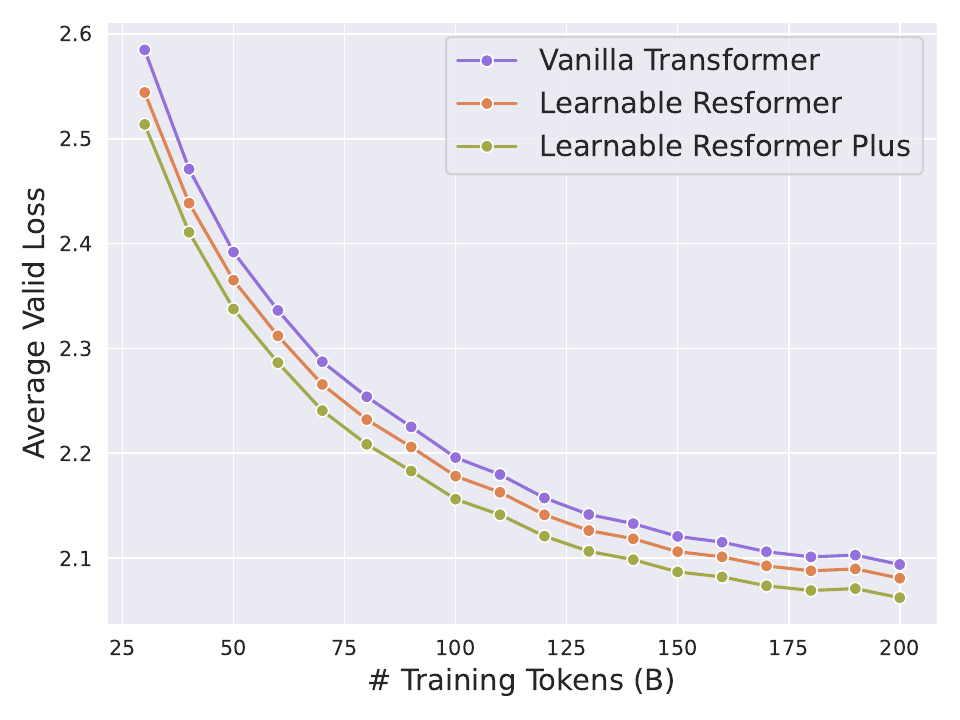}
      \end{subfigure}
    \caption{
       (Left) Validation loss as model size scales from 82M to 468M parameters on 20B tokens. (Medium) Validation loss for the 468M parameter model evaluated every 2B tokens. ResFormer achieves approximately 16.1\%-20.3\% reduction in both model parameters and training data. (Right) Validation loss for the 1.6B parameter model evaluated every 10B tokens.}
    \label{fig:scaling_loss}
\end{figure*}
\begin{table*}[h]
\centering
\scriptsize
\begin{tabular}{cccccccccccc}
\toprule
Model & Wiki. & LMB. & Arc-c & Arc-e & BoolQ & Hella. & LMB. & Obqa. & Wino. & PiQA & Avg. \\
 & PPL & PPL & ACC & ACC & ACC & ACC & ACC & ACC & ACC & ACC & ACC \\
\midrule
Transformer & 24.8 & 33.4 & 18.8 & 49.6 & 57.5 & 31.1 & 34.1 & 17.6 & 49.8 & 66.1 & 40.6 \\
NeuTRENO & 24.3 & 36.3 & 20.7 & 48.8 & 59.2 & 31.6 & 34.2 & \textbf{19.0} & \textbf{51.8} & 65.9 & 41.4 \\
DenseFormer & 24.0 & \textbf{28.0} & 20.1 & 48.7 & 56.5 & 32.1 & \textbf{36.6} & 17.6 & 49.3 & 65.6 & 40.8 \\
Identity ResFormer & 23.8 & 32.9 & 20.2 & 49.0 & 59.0 & 32.1 & 36.0 & 16.8 & 51.2 & 66.1 & 41.3 \\
Learnable ResFormer & 23.7 & 32.5 & \textbf{21.2} & \textbf{50.3} & \textbf{60.9} & 32.3 & 36.3 & 18.8 & 51.2 & 67.0 & \textbf{42.3} \\
Learnable ResFormer plus & \textbf{23.2} & 31.4 & \textbf{21.2} & 49.7 & 60.6 & \textbf{32.4} & 36.0 & 17.8 & 51.1 & \textbf{67.5} & 42.0 \\
\bottomrule
\end{tabular}
\caption{Downstream evaluation of different models with 468M parameters trained on 20B tokens.}
\label{tab:model_comparison}
\end{table*}

Beyond ResFormer, SVFormer adopts standard attention in the first layer and obtain the attention output $\mathbf{U_n}$ for $n$-th layer when $n \geq 2$ through $\mathbf{U}_{n} = \mathbf{A}_{n} {\mathbf{V}_{1}}$, where $\mathbf{A}_{n}$ is the attention matrix of $n$-th layer. Its main advantage is that it only requires computing and storing the value vectors for the first layer, saving nearly half of the $KV$ cache during inference. Similar methods like CLA reduce $KV$ cache by sharing both of the key and value vectors every two layers. However, the results in Table ~\ref{tab:CLA_ablation_kv} show that sharing values has less negative impact compared with sharing keys. 

\section{Experiments}
\subsection{Setting}
\label{sec:training_details}
\paragraph{Training Details}Following \cite{brandon2024reducing}, we choose the Llama-like architecture and SlimPajama \citep{cerebras2023slimpajama} data for main experiments. Specifically, the architecture includes pre-normalization, SwiGLU activations \citep{shazeer2020glu}, rotary position embedding \citep{su2024roformer}, and no dropout. For SlimPajama, we randomly sample nearly 20B tokens based on the original data distribution of seven domains during training and adopt tokenizer used for ``RedPajama-INCITE-7B-Base". See Table \ref{tab:slimpajama-data-proportion} in Appendix for data details.

Unless otherwise noted, we train all models using AdamW optimizer with 0.1 weight decay, $\beta_1=0.9$, $\beta_2=0.95$ and the max grad norm 1.0.
The batch size is set to be around 2M tokens \citep{zhang2024tinyllama} with a sequence length of 2,048 and the total steps is fixed 10,000 steps \citep{kaplan2020scaling}. We adopt linear learning rate warmup for the first 1,200 steps with the initial learning rate and the peak learning rate to be 1e-7 and 6e-4 respectively. The cosine decay schedule gradually decays to 10\% of the peak learning rate by the end of training \citep{zhou2024lima, wei2023skywork}. The detailed hyperparameters for models of various sizes and different training sequence lengths in Appendix~\ref{sec:training_details_appendix}. Moreover, All models are trained with 8 Nvidia A100 80G GPUs using mixed-precision training in FP16. We adopt deepspeed zero-2 optimizer and flash attention mechanism.

\paragraph{Scaling Details} We conduct scaling experiments on a 1.6B parameter model using approximately 200B tokens of internal pre-training data. Training is performed with a sequence length of 8,192, learning rate of 1.5e-4, and batch size of 2M tokens, requiring about 3,584 H800 GPU hours per run.

\paragraph{Evaluation}
For model evaluation and comparison, we primarily utilized the average validation loss across seven domains, computed on the entire SlimPajama validation split. Additionally, we randomly selected a fixed set of 1,000 sample sequences for subsequent visualization analysis.

We also compare different models on several classical reasoning tasks following \citep{zhang2024tinyllama} in a zero-shot way. The tasks include Hellaswag (Hella.) \citep{zeller2019hellaswag} Openbookqa (Obqa.) \citep{mihaylov2018suit}, WinoGrande (Wino.) \citep{sakaguchi2019winogrande}, ARC-Easy (Arc-e) and ARC-Challenge (Arc-c) \citep{clark2018think} and PiQA \citep{bisk2020piqa}. In addition to accuracy on reasoning tasks, we also report the perplexity (PPL) on Wikitext (Wiki.) and LAMBADA (LMB.).

\subsection{ResFormer vs. NeuTRENO, DenseFormer}
We analyze how different models scale with model size and data size under similar experimental settings. We train models with 82M, 180M, 320M, and 468M parameters on 20B tokens and evaluate on a validation set. Fig.\ref{fig:scaling_loss} (Left) shows ResFormer achieves equivalent validation loss to Transformer while using 16.11\% fewer parameters. We also evaluate 468M models every 2B tokens, finding ResFormer requires 20.3\% fewer training tokens to match Transformer's loss (Fig.\ref{fig:scaling_loss} (Medium)). All ResFormer variants demonstrate superior scaling compared to NeuTRENO and DenseFormer. Table~\ref{tab:model_comparison} confirms ResFormer outperforms other models on downstream tasks, with Learnable ResFormer achieving 1.7 point average accuracy improvement over vanilla Transformer.

To validate the performance improvement of value residual at larger training scales, we conduct additional experiments following the scaling details in Sec.\ref{sec:training_details}. As shown in Fig.\ref{fig:scaling_loss} (Right), value residual consistently improves performance throughout training. 

\begin{figure}[!h]
    \centering
    \includegraphics[width=0.7\linewidth]{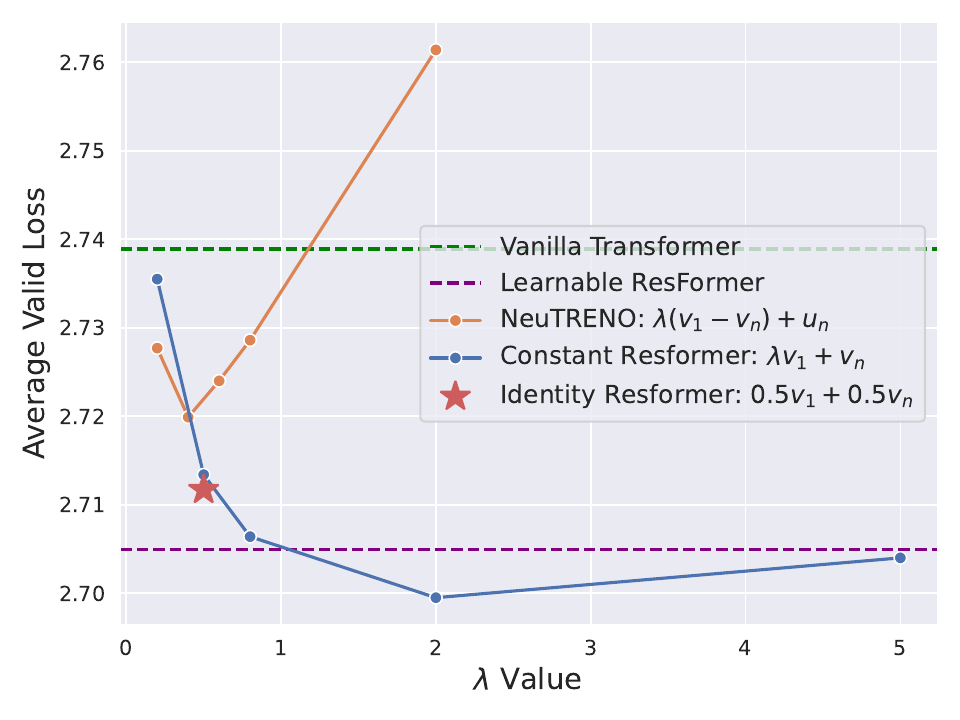}
    \vspace{-3mm}
    \caption{
       The impact of varying $\bm\lambda$ values on 82M 8-layer Constant-ResFormer and NeuTRENO.}
    \label{fig:value_residual_which_lambda}
\end{figure}

Both Constant-ResFormer and NeuTRENO rely on predetermined $\bm\lambda$ constants. Fig.~\ref{fig:value_residual_which_lambda} shows the performance curves of these models against varying $\bm\lambda$. Results indicate that Constant-ResFormer significantly outperforms NeuTRENO and demonstrates greater robustness across a wider range of $\bm\lambda$ values, achieving optimal performance at $\bm\lambda=2$.
\begin{table}[!ht]
\centering
\scriptsize
\renewcommand{\arraystretch}{1.5}
\setlength{\tabcolsep}{3pt}
\begin{tabular}{ccc}
\toprule
Model & Initial Form  & Loss \\ \midrule
\rowcolor{gray!30}\multicolumn{3}{c}{Baselines} \\
\hline
 Vanilla Transformer &\scriptsize- & 2.739 \\
\hdashlineself
 $\text{DenseFormer}^*$ &\scriptsize$\mathbf{H}_{n}^{\prime}\!=\!\textcolor{red}{1\times}\mathbf{H}_{n}+\sum_{i = 0}^{n-1}\textcolor{red}{0\times}\mathbf{H}_{i}$ & 2.722 \\ 
\hdashlineself
 NeuTRENO 
&\scriptsize$\mathbf{U}_{n}^{\prime}=\textcolor{red}{0.4}(\mathbf{V}_{1}-\mathbf{V}_{n})+\mathbf{U}_{n}$ & 2.72 \\
\hdashlineself
\hline
\rowcolor{gray!30}\multicolumn{3}{c}{ResFormer} \\
\hline
Identity-ResFormer &\scriptsize$\mathbf{V}_{n}^{\prime} = \textcolor{red}{0.5}\mathbf{V}_{1}+\textcolor{red}{0.5}\mathbf{V}_{n}$ & 2.712 \\
\hdashlineself
$\text{Dense-ResFormer}^*$ &\scriptsize$\mathbf{V}_{n}^{\prime} = \sum_{i = 1}^{n}\textcolor{red}{1\times}\mathbf{V}_{i}$ & 2.709 \\
\hdashlineself
$\text{Learnable-ResFormer}^*$ &\scriptsize$\mathbf{V}_{n}^{\prime} = \textcolor{red}{0.5}\mathbf{V}_{1}+\textcolor{red}{0.5}\mathbf{V}_{n}$ & 2.705 \\
\hdashlineself
Constant-ResFormer &\scriptsize$\mathbf{V}_{n}^{\prime} = \textcolor{red}{2}\mathbf{V}_{1}+\mathbf{V}_{n}$ & 2.7 \\
\hdashlineself
Sparse-ResFormer &\scriptsize$\left\{
  \begin{array}{l}
    \mathbf{V}_{n}^{\prime} = \mathbf{V}_{n},1 \leq n \leq 5 \\
    \mathbf{V}_{n}^{\prime} = \mathbf{V}_{1}, 6 \leq n \leq 8
  \end{array}
  \right.$ & 2.696 \\
\hdashlineself
Sparse-ResFormer &\scriptsize$\left\{
  \begin{array}{l}
    \mathbf{V}_{n}^{\prime} = \mathbf{V}_{n},1 \leq n \leq 5 \\
    \mathbf{V}_{n}^{\prime} = \textcolor{red}{5}\mathbf{V}_{1}+\mathbf{V}_{n}, \\
    \qquad\ \quad\qquad6 \leq n \leq 8
  \end{array}
  \right.$ & 2.687 \\
\hdashlineself
$\text{ResFormer-Plus}^*$ & See Sec.\ref{sec:resformer_formulation} & 2.681 \\
\bottomrule
\end{tabular}%
\caption{Average valid loss for 8-layer, 82M-parameter models. ``Initial form" shows deviations from vanilla transformer. \textcolor{red}{Red} numbers are the $\bm\lambda$ values from Eqn.~\ref{eqn:NeuTRENO}, Eqn.~\ref{eqn:DenseFormer}, and Eqn.~\ref{eqn:resformer}. For models marked with ``*" , $\bm\lambda$ is learnable, and the red numbers indicate the initial value; otherwise, red numbers are fixed constants.}
\label{tab:resformer_vs_denseformer}
\end{table}

Furthermore, we test the performance of other varaints of ResFormer mentioned in Sec.\ref{sec:resformer_formulation}. The $\bm\lambda$ values for Constant-ResFormer and Sparse-ResFormer were optimized through multiple experiments. All ResFormer variants, including the simplest Identity-ResFormer, show significant performance improvements. However, manually-tuned Sparse-ResFormer and Learnable-ResFormer-Plus outperform the standard Learnable-ResFormer. It demonstrates the challenge to pre-determine the optimal layers for $\mathbf{V}_{1}$ connections and their corresponding $\bm\lambda_{1}$ values in more general scenarios. Interestingly, the third to last row shows that Sparse-ResFormer achieved better performance despite having three fewer $\mathbf{W}^{\mathbf{V}}$.

\subsection{Truly Better or Just Faster?}
\begin{figure}[!h]
    \centering
      \begin{subfigure}{0.49\linewidth}
        \centering
        \includegraphics[width=\linewidth]{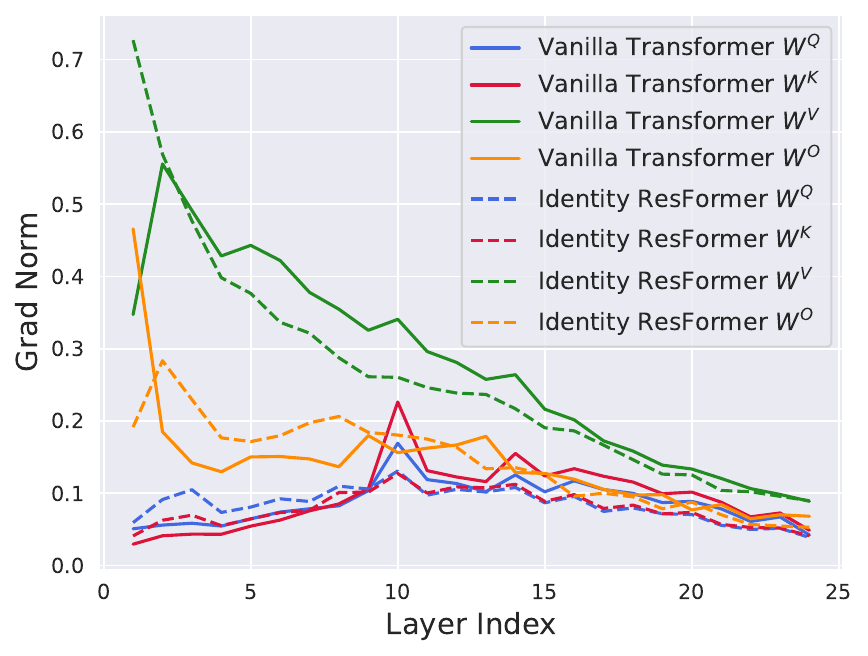}
      \end{subfigure}
      \begin{subfigure}{0.49\linewidth}
        \centering
        \includegraphics[width=\linewidth]{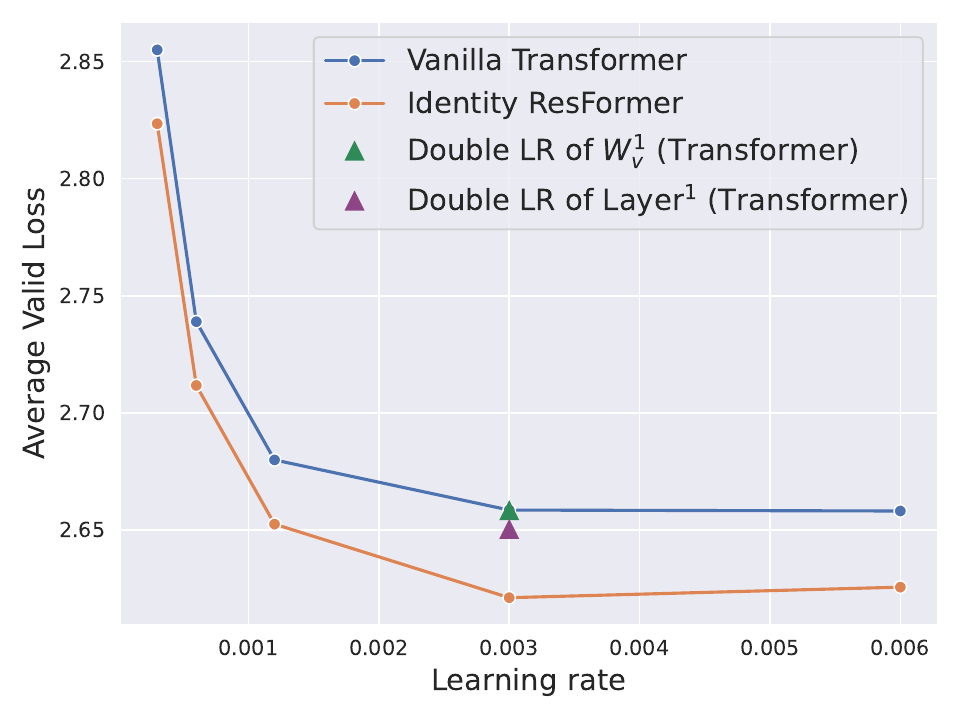}
      \end{subfigure}
    \vspace{-5mm}
    \caption{
       (Left) Average gradient norms of model outputs with respect to parameter matrices across different layers in Transformer and ResFormer. (Right) Comparison of Transformer and ResFormer performance across various learning rates during training.}
    \label{fig:resformer_lr}
\end{figure}
To verify that ResFormer's performance improvements are not solely due to accelerated training from its shortcuts, we examined model performance across different learning rates. We compared Identity ResFormer and vanilla Transformer under five learning rate settings. As shown in Fig.~\ref{fig:resformer_lr} (Right), both models achieved optimal results around a learning rate of 0.003, with Identity ResFormer significantly outperforming vanilla Transformer across all rates.

Analysis of the grad norm for the four parameter matrices ($\mathbf{W}^\mathbf{Q}$, $\mathbf{W}^\mathbf{K}$, $\mathbf{W}^\mathbf{V}$, $\mathbf{W}^\mathbf{O}$) in each layer's attention module revealed that Identity ResFormer's output had approximately twice the grad norm for $\mathbf{W}^\mathbf{V}_1$ and half for $\mathbf{W}^\mathbf{O}_1$ in the first layer compared to vanilla Transformer. This indicates that a portion of the gradient originally propagated to $\mathbf{V}_1$ through $\mathbf{H}_1$ is now transmitted via the value residual directly for Identity ResFormer.

In this way, we conducted the other two ablation experiments on vanilla Transformer: doubling the learning rate for only the first layer, and doubling it exclusively for $\mathbf{W}^\mathbf{V}_1$ in the first layer. Neither modification yielded significant improvements. This further demonstrates that the performance improvements brought by ResFormer are unrelated to the changes in gradient magnitude.
\subsection{Ablation Study of Value Residual}
\paragraph{Where from, where to?}
\begin{figure}[!h]
    \centering
      \begin{subfigure}{0.49\linewidth}
        \centering
        \includegraphics[width=\linewidth]{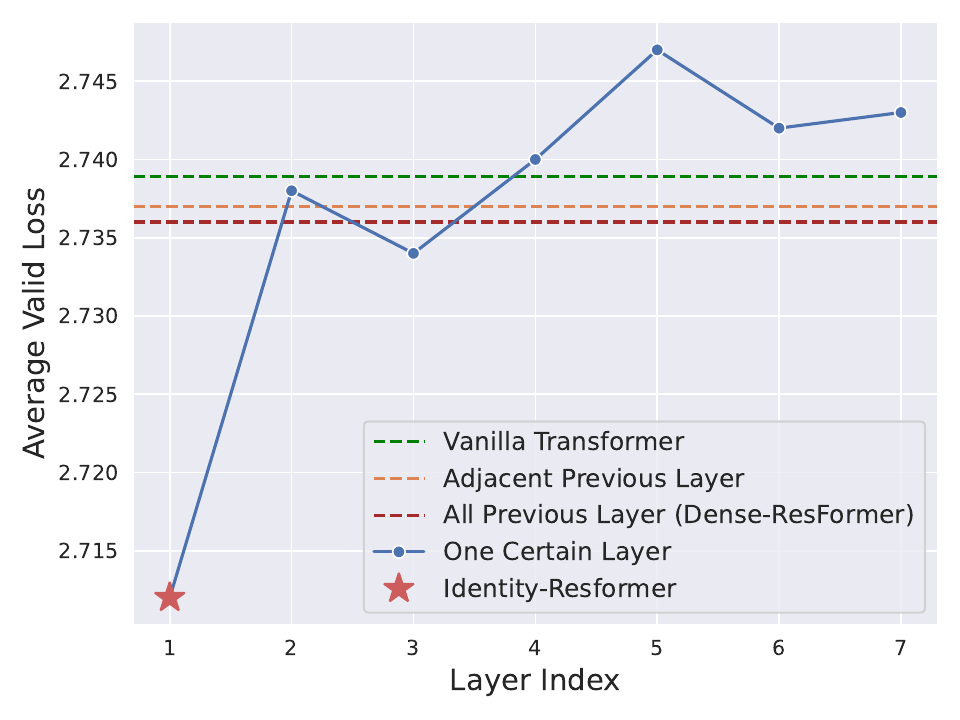}
      \end{subfigure}
      \begin{subfigure}{0.49\linewidth}
        \centering
        \includegraphics[width=\linewidth]{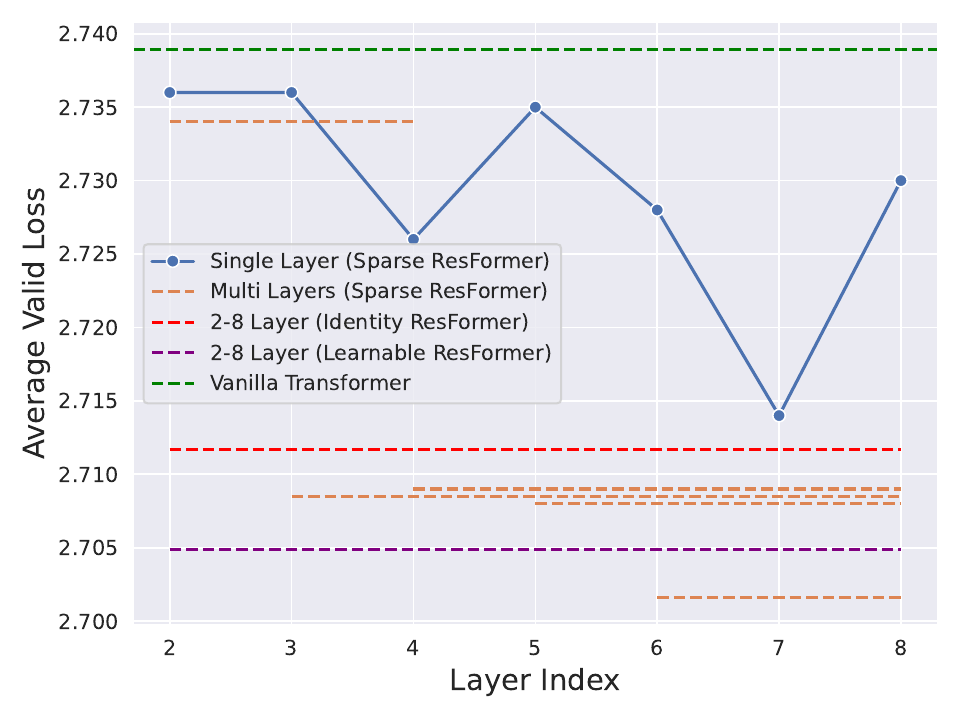}
      \end{subfigure}
    \vspace{-5mm}
    \caption{
       (Left) Impact of value skip connections source from different layers on model performance, where all connections are identity connections and $\bm\lambda=1$ in Dense-ResFormer. (Right) Average validation loss of various Sparse-ResFormer configurations, which retain only single or multiple skip connections from $\mathbf{V}_{1}$.}
    \label{fig:value_residual_target_source}
\end{figure}
We analyzed which value skip-connections are necessary for the vanilla transformer. For an 8-layer transformer, we added various pre-defined value skip-connections (with constant $\bm{\lambda}$) and evaluated the resulting validation loss. As shown in Fig.~\ref{fig:value_residual_target_source} (Left), we first examined the impact of skip-connections from different sources. Our findings indicate that only skip-connections originating from the first layer's value ($\mathbf{V}_{1}$) yield significant performance improvements. Skip-connections from the second layer's value ($\mathbf{V}_{2}$) offer no significant benefit to subsequent layers. Skip-connections from later layers, occurring only in the final few layers, even lead to performance degradation. Both of the two special cases in Fig.~\ref{fig:value_residual_target_source} (Left) include $\mathbf{V}_{1}$ skip-connections. However, when these connections occur only between adjacent layers, the information in $\mathbf{V}_{1}$ fails to effectively reach the final layers. Conversely, dense value skip-connections dilute the impact of $\mathbf{V}_{1}$ with information from other sources.

Furthermore, we investigated spare ResFormer, a variant of identity ResFormer where the value residual connection $\mathbf{V}_{n}^{\prime} = {0.5}\mathbf{V}_{1}+{0.5}\mathbf{V}_{n}$ is applied selectively to specific layers. As shown in Fig.~\ref{fig:value_residual_target_source} (Right), for an 8-layer model, when limited to a single layer, applying the residual connection to the 7th layer yields the most significant improvement. When applied to multiple layers, the greatest benefit is observed when incorporating layers 6 to 8. Extending the residual connection to earlier layers, such as the 5th, diminishes the overall effect. It suggests that the model's final few layers benefit most from the first layer's value information.

\begin{figure}[!h]
    \centering
      \begin{subfigure}{0.49\linewidth}
        \centering
        \includegraphics[width=\linewidth]{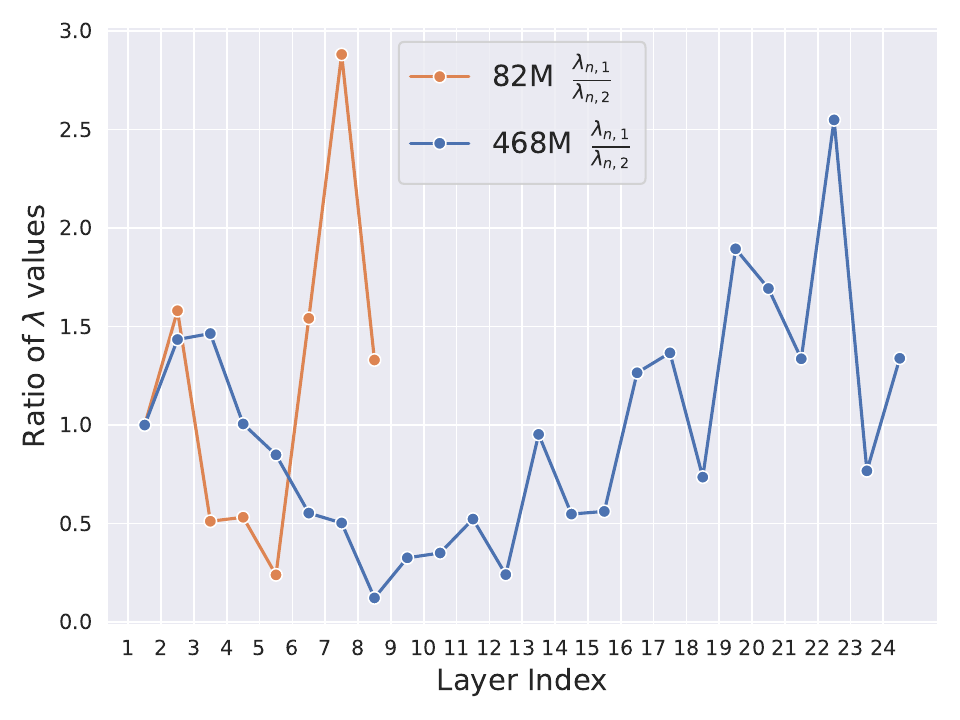}
      \end{subfigure}
      \begin{subfigure}{0.49\linewidth}
        \centering
        \includegraphics[width=\linewidth]{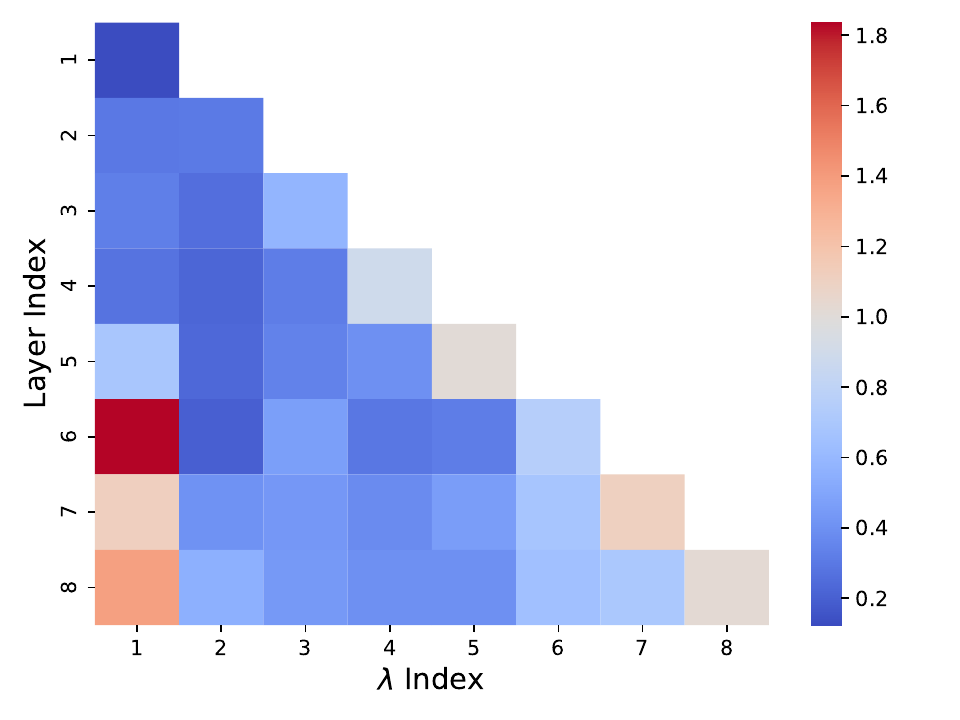}
      \end{subfigure}
    \vspace{-5mm}
    \caption{
       (Left) Visualization of $\bm\lambda_1/\bm\lambda_2$ across different layers in the 82M and 468M Learnable-ResFormer. (Right) Heatmap visualization of learned $\bm\lambda$ across different layers in the 468M Dense-Learnable-ResFormer.}
    \label{fig:lambda_vis}
\end{figure}
We further trained 8-layer and 24-layer Learnable-ResFormers, as well as an 8-layer Learnable-Dense-ResFormer, and visualized the learned $\bm{\lambda}$ values. As shown in Fig.~\ref{fig:lambda_vis}, the later layers tend to require more value residual connections from $\mathbf{V}_{1}$, which aligns with the findings in Fig.~\ref{fig:value_residual_target_source}. Fortunately, the Learnable-ResFormer can, to some extent, identify similar sparse residual patterns to those of the best performing Sparse-ResFormer in Table~\ref{tab:resformer_vs_denseformer}. Notably, the Learnable-Dense-ResFormer learns value residual patterns that closely resemble those of the Learnable-ResFormer.

\paragraph{Why needed beyond hidden residual?}
\begin{table}[!h]
    \centering
    \resizebox{0.99\linewidth}{!}{
    \begin{tabular}{ccc}
    \toprule
    \makecell{Residual\\Type} & Initial Form & \makecell{Valid\\Loss} \\
    \midrule
    - & - & 2.7389 \\
    value & $\mathbf{V}_{n}^{\prime} = \textcolor{red}{0.5}\mathbf{V}_{1}+\textcolor{red}{0.5}\mathbf{V}_{n}$ & 2.705 \\
    hidden & $\mathbf{H}_{n}^{\prime} = \textcolor{red}{0.5}\mathbf{H}_{0}+\textcolor{red}{0.5}\mathbf{H}_{n}$ & 2.781 \\
    value & $\mathbf{V}_{n}^{\prime} = 
    \textcolor{red}{0\times}\mathbf{V}_{1}+\textcolor{red}{1}\mathbf{V}_{n}$ & 2.73 \\
    hidden & $\mathbf{H}_{n}^{\prime} = 
    \textcolor{red}{0\times}\mathbf{H}_{0}+\textcolor{red}{1}\mathbf{H}_{n}$ & 2.722 \\
    \bottomrule
    \end{tabular}}
    \caption{Comparison of additional value residual (to $\mathbf{V}_{1}$) and hidden residual (to $\mathbf{H}_{0}$) connections against the default hidden residual, under various $\mathbf{\lambda}$ initializations. Trainable $\mathbf{\lambda}$ parameters are highlighted in \textcolor{red}{red}.}
    \label{tab:value_residual_vs_hidden}
\end{table}
Our experiments revealed that $\mathbf{V}_{1}$ information provides additional benefits to later network layers, despite both $\mathbf{H}_{0}$ and $\mathbf{V}_{1}$ containing initial, unfused token information. $\mathbf{H}_{0}$ is propagated through default hidden residual connections, but it may be diluted by subsequent information, hindering its effective utilization in later layers. To test this hypothesis, we introduced an additional skip connection to $\mathbf{H}_{0}$ : $\mathbf{H}_{n}^{\prime} = {\mathbf{\lambda}_0}\mathbf{H}_{1}+{\mathbf{\lambda}_2}\mathbf{H}_{n}$, where $\mathbf{\lambda}$ is learnable. We conducted experiments with two $\mathbf{\lambda}$ initialization settings and compared them to value residual.

Results showed that when ${\mathbf{\lambda}_1}={\mathbf{\lambda}_2}$ initially, the extra hidden residual had adverse effects. However, initializing ${\mathbf{\lambda}_1}=0$ yielded some improvements, suggesting possible dilution of $\mathbf{H}_{0}$ information. Nevertheless, these gains were smaller than those from value residual connections, which consistently outperformed vanilla transformers across different initializations. Actually, the connection $\mathbf{H}_{0}$ : $\mathbf{H}_{n}^{\prime} = {\mathbf{\lambda}_1}\mathbf{H}_{1}+{\mathbf{\lambda}_2}\mathbf{H}_{n}$, is similar to applying residuals to queries, keys, and values at the same time, may disrupt attention distributions and hinder higher-level semantic information fusing. The reduction performance brought by identity residuals of queries or keys shown in Table~\ref{tab:residual_type} can support it.
\paragraph{Why $\mathbf{V}_1$ instead of $\mathbf{V}_2$?}
\begin{table}[htbp]
    \centering
    \resizebox{0.8\linewidth}{!}{
    \begin{tabular}{ccc}
    \toprule
    \makecell{Hidden Residual\\Starts Place} & \makecell{Value Residual\\Target Place} & \makecell{Valid\\Loss} \\
    \midrule
    $\mathbf{H_0}$ (Default) & - & 2.739 \\
    $\mathbf{H_0}$ (Default) & $\mathbf{V_1}$ & 2.712 \\
    $\mathbf{H_0}$ (Default) & $\mathbf{V_2}$ & 2.738 \\
    \rowcolor{gray!30}$\mathbf{H_1}$ & - & 2.78 \\
    \rowcolor{gray!30}$\mathbf{H_1}$ & $\mathbf{V_2}$ & 2.78 \\
    \rowcolor{gray!30}$\mathbf{H_2}$ & - & 2.82 \\
    \rowcolor{gray!30}$\mathbf{H_2}$ & $\mathbf{V_2}$ & 2.787 \\
    \rowcolor{lightgray!30}$\mathbf{H_2}$ & - & 2.82 \\
    \rowcolor{lightgray!30}$\mathbf{H_2}$ & $\mathbf{V_3}$ & 2.833 \\
    \rowcolor{lightgray!30}$\mathbf{H_3}$ & - & 3.057 \\
    \rowcolor{lightgray!30}$\mathbf{H_3}$ & $\mathbf{V_3}$ & 2.883 \\
    \bottomrule
    \end{tabular}}
    \caption{Comparison of performance across different value residual target and varying hidden residual start settings. ``value residual target place" $\mathbf{V}_i$ indicates the earliest value accessible to subsequent layers, while ``hidden residual starts place" denotes the earliest hidden state available, without prior residual connections.}
    \label{tab:value_residual_different_resnet}
\end{table}

\begin{figure}[!h]
    \centering
      \begin{subfigure}{0.49\linewidth}
        \centering
        \includegraphics[width=0.9\linewidth]{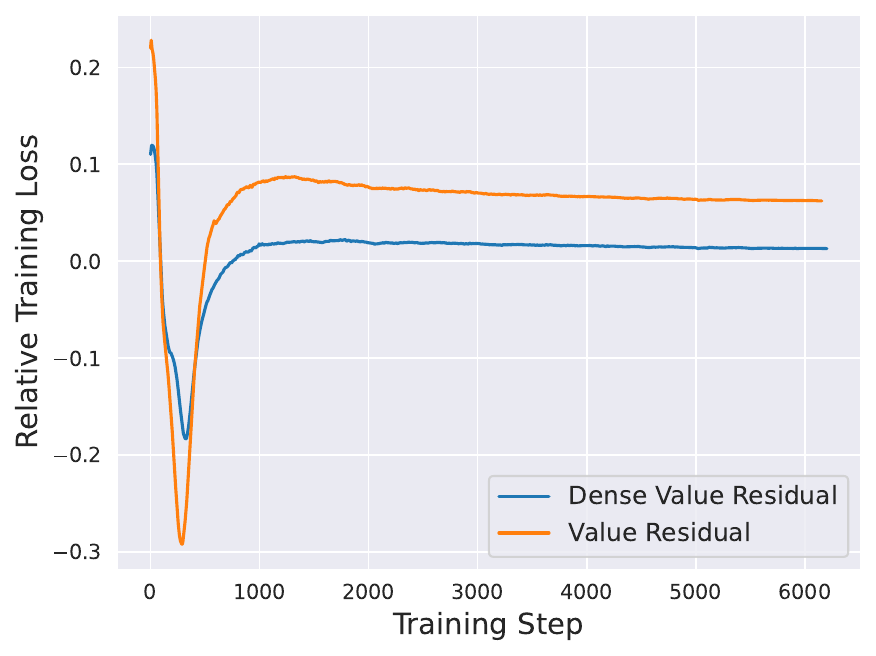}
      \end{subfigure}
      \begin{subfigure}{0.49\linewidth}
        \centering
        \includegraphics[width=\linewidth]{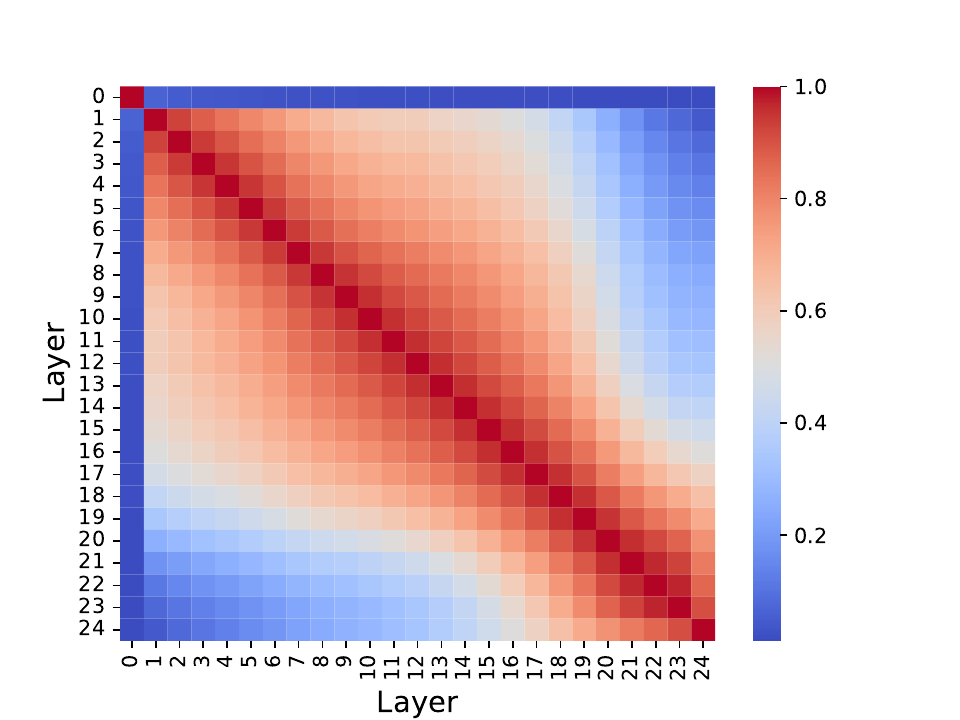}
      \end{subfigure}
    \caption{
       (Left) The relative training loss curve between different cross layer residual and vanilla hidden residual. (Right) Layer-to-layer hidden states similarity.}
    \label{fig:cross_layer_residual}
\vspace{-4mm}
\end{figure}
In Fig.~\ref{fig:value_residual_target_source} (Left), connections to $\mathbf{V}_1$ show significant improvement, while those to $\mathbf{V}_2$ yield minimal gains. This likely occurs because the original hidden residual propagates information from $\mathbf{H}_1$ to the network ($\mathbf{V}_2=\mathbf{H}_1\mathbf{W}^{\mathbf{V}}_2$). To verify, we adjusted residual connections, introducing them at different points. For example, starting from $\mathbf{H}_1$, we use $\mathbf{H}_1=Layer_1(\mathbf{H}_0)$ instead of $\mathbf{H}_1=Layer_1(\mathbf{H}_0)+\mathbf{H}_0$.

Table~\ref{tab:value_residual_different_resnet} results show that when residual connections begin from $\mathbf{H}_0$ or $\mathbf{H}_1$, allowing $\mathbf{H}_2$ and subsequent layers access to $\mathbf{H}_1$, $\mathbf{V}_2+\mathbf{V}_n$ offers no improvement. However, starting from $\mathbf{H}_2$, skip connections from $\mathbf{V}_2$ provide substantial benefits. Regarding the disparity in information propagation between $\mathbf{V}_2$ (via $\mathbf{H}_1$) and $\mathbf{V}_1$ (via $\mathbf{H}_0$), we posit that after the first layer's integration, $\mathbf{H}_1$ contains higher-level semantic information more similar to subsequent hidden states, see Fig.~\ref{fig:cross_layer_residual} (Right). This may ensure that the attention distribution remains relatively undisturbed when connecting to $\mathbf{H}_1$. Besides, Fig.~\ref{fig:cross_layer_residual} (Left) shows that Dense-ResFormer performs better than ResFormer when there is no cross layer hidden residual.
\paragraph{Superior to other residual}

\begin{table}[htbp]
\small
\centering
\begin{minipage}{0.47\linewidth}
    \centering
    \resizebox{\linewidth}{!}{
    \begin{tabular}{cc}
    \toprule
    Residual Type & \makecell{Valid\\Loss} \\
    \midrule
    - & 2.739 \\
    Query & 2.742 \\
    Key & 2.746 \\
    Attention & 2.757 \\
    Value & 2.712 \\ 
    \bottomrule
    \end{tabular}}
    \caption{The impact of various residual types, where all residual connections adopt a form similar to $\mathbf{V}_{n}^{\prime}\!=\!{0.5}\mathbf{V}_{1}+{0.5}\mathbf{V}_{n}$.}
    \label{tab:residual_type}
\end{minipage}%
\hspace{0.03\linewidth}
\begin{minipage}{0.47\linewidth}
    \centering
    \resizebox{\linewidth}{!}{
    \begin{tabular}{cc}
    \toprule
    Residual Mapping & \makecell{Valid\\Loss} \\
    \midrule
    - & 2.739 \\
    Identity Mapping & 3.137 \\ 
    \makecell{Cross Layer\\Attention} & 2.729 \\
    Current Attention & 2.712 \\
    \bottomrule
    \end{tabular}}
    \caption{Comparison of different mapping matrices when adding $\mathbf{V}_1$ to $\mathbf{U}_n$, with ``Current Attention" corresponding to Identity-ResFormer.}
    \label{tab:residual_mapping}
\end{minipage}
\end{table}
For vanilla transformers, to better propagate information from the first layer, new residual connections can be introduced at various points in addition to the existing hidden residual: query states $\mathbf{Q}$, key states $\mathbf{K}$, value states $\mathbf{V}$, and post-softmax attention matrix $\mathbf{A}$. Results in Table~\ref{tab:residual_type} indicate that only the value residual connection improves performance. When connecting $\mathbf{V}_{1}$ and $\mathbf{V}_{n}$, three approaches free of extra parameters are possible: (1) the proposed residual connection, directly summing the two and then sharing an attention matrix; (2) cross layer attention (\small${\operatorname{Softmax}\Big({\mathbf{Q}_{n}\operatorname{Concat}(\mathbf{K}_{n},\mathbf{K}_{1})^T}\Big) \operatorname{Concat}(\mathbf{V}_{n},\mathbf{V}_{1})}$\normalsize), recomputing an attention matrix for $\mathbf{V}_{1}$ based on $\mathbf{K}_{1}$ and $\mathbf{Q}_{n}$; and (3) directly adding $\mathbf{V}_{1}$ to $\mathbf{U}_{n}$ in Eqn.~\ref{eqn:attention}, equivalent to using an identity mapping as $\mathbf{V}_{1}$'s attention matrix in layer $N$. The second approach significantly increases computational cost. Results in Table~\ref{tab:residual_mapping} demonstrate that sharing the attention matrix yields the best performance.

\subsection{Post-Analysis of ResFormer}
\begin{figure}[!h]
    \centering
      \begin{subfigure}{0.49\linewidth}
        \centering
        \includegraphics[width=\linewidth]{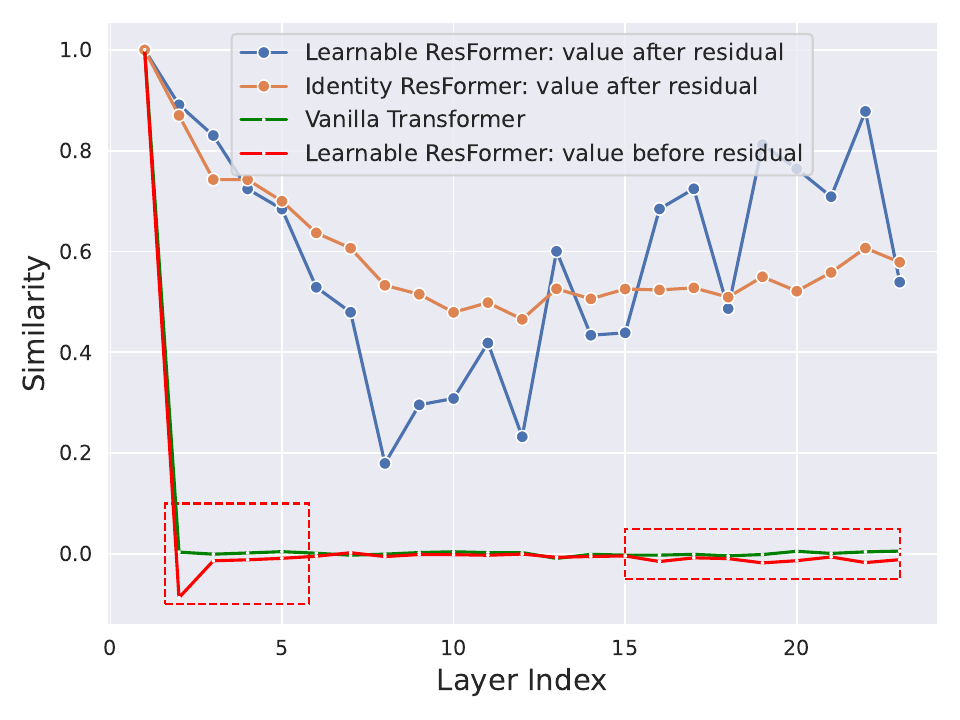}
      \end{subfigure}
      \begin{subfigure}{0.49\linewidth}
        \centering
        \includegraphics[width=\linewidth]{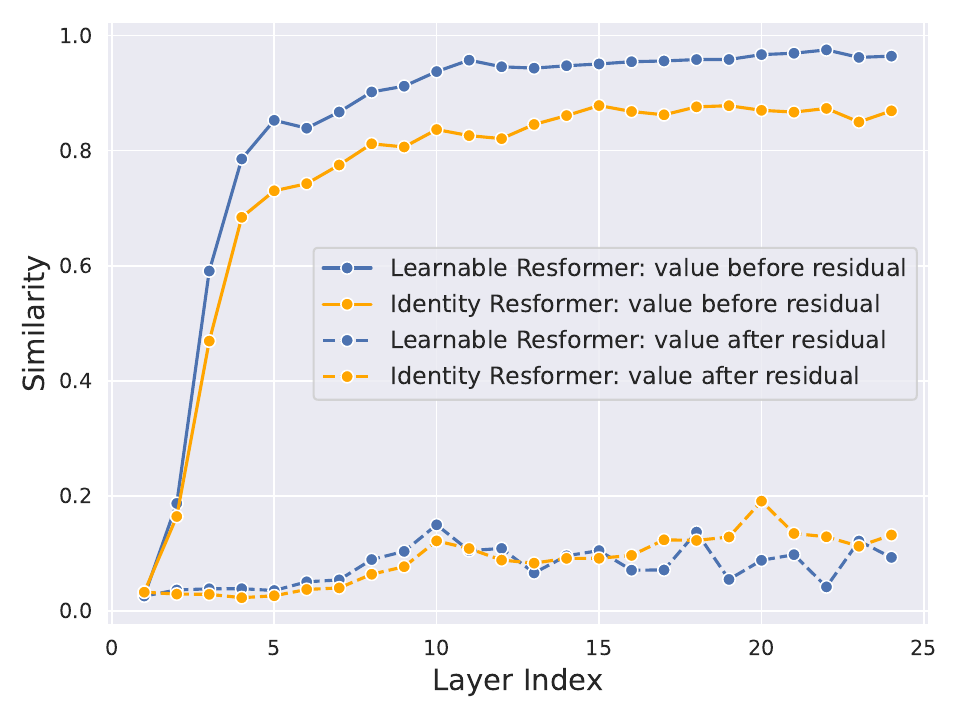}
      \end{subfigure}
    \caption{
       (Left) Similarity between first layer values and other layers' values. (Right) Token-to-token similarity across sequence for value states at different place.}
    \label{fig:value_similarity}
\vspace{-5mm}
\end{figure}
\paragraph{How value residual works?}We performed post-analysis on trained ResFormer and vanilla Transformer models to understand value residual learning. Fig.~\ref{fig:value_similarity} (Left) shows cosine similarities between value states at different layers and the first layer, averaged across token positions. For ResFormer, we calculated this before and after applying value residual. Results show that in vanilla Transformers, the first layer's value has low similarity with other layers. In contrast, ResFormer maintains high similarity between the first layer's value and the post-residual values in subsequent layers due to value residual connections. Notably, in layers where ResFormer relies more heavily on the first layer's value (see Fig.~\ref{fig:lambda_vis}), the pre-residual value exhibits lower similarity with the first layer's value, indicating that $\mathbf{W}^\mathbf{V}$ in these layers is learning the value residual.

For ResFormer, we also examined the average pairwise similarity between tokens' values before and after the residual connection. The results Fig.~\ref{fig:value_similarity} (Right) reveal that with value residual connections, the learned values (before the value residual) from each layer become increasingly similar as the network deepens. We hypothesize that this is because, given the default hidden residual and value residual, each layer learns a $\Delta \mathbf{V}$, with the magnitude of necessary adjustments decreasing in later layers. This phenomenon is unique to ResFormer and not observed in vanilla Transformers.
\begin{figure}[!h]
    \centering
      \begin{subfigure}{0.49\linewidth}
        \centering
        \includegraphics[width=\linewidth]{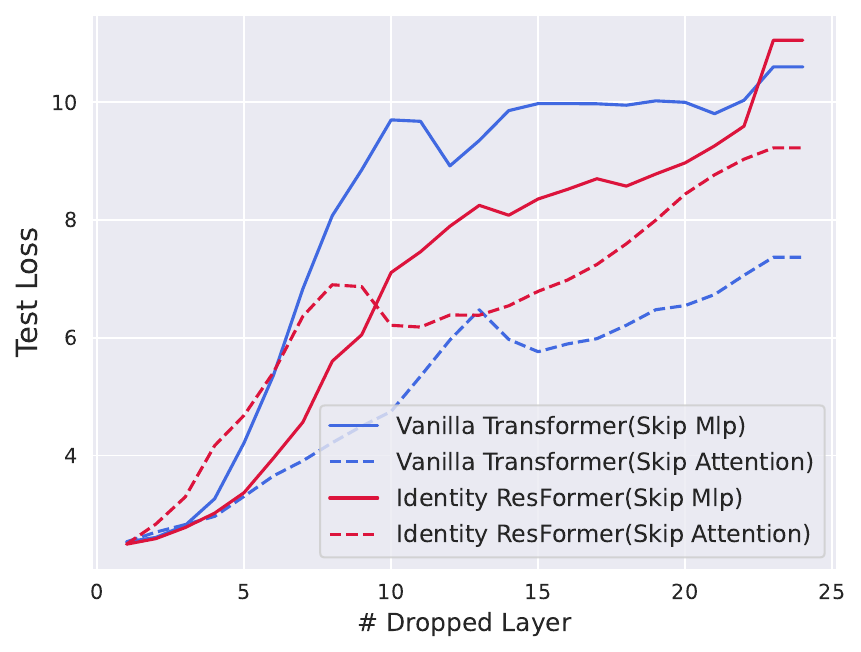}
      \end{subfigure}
      \begin{subfigure}{0.49\linewidth}
        \centering
        \includegraphics[width=\linewidth]{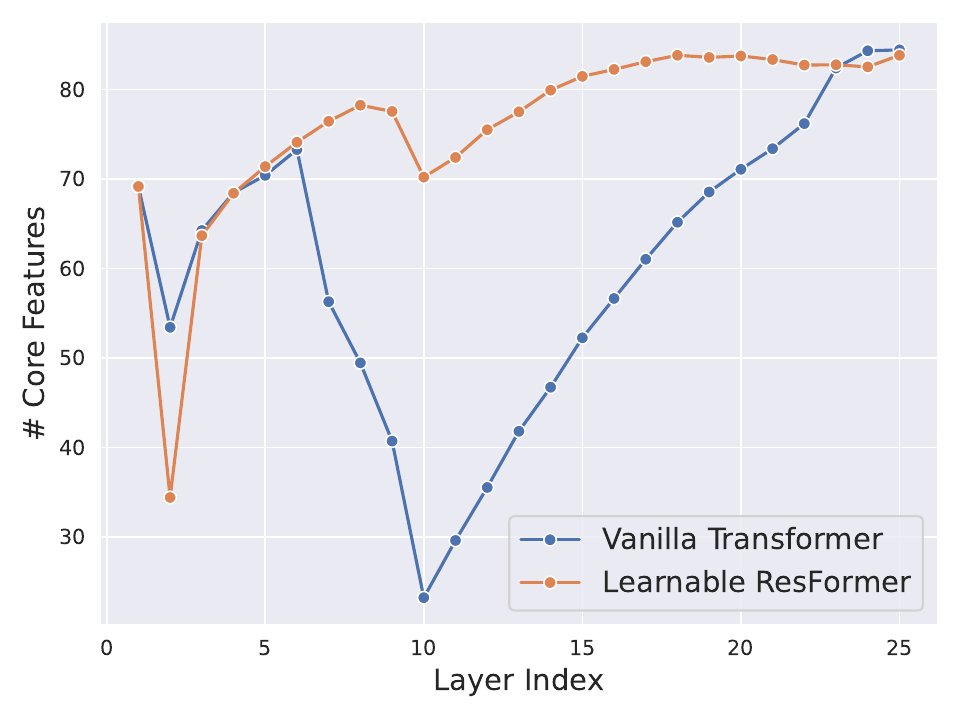}
      \end{subfigure}
    \caption{
       (Left) The change in test loss as model modules are progressively removed, starting from the back to front while keeping the first layer intact.  (Right) The number of core features in each layer's hidden state after PCA dimensionality reduction, where core features represent the minimum number of principal components required to explain $99\%$ of the variance.}
    \label{fig:pca_skip_analysis}
\end{figure}
\paragraph{Representation and Module Analysis} We analyzed the overall network changes, focusing on the hidden state representation capabilities and the contributions of different modules. \cite{Tyukin2024attention} suggests that removing $\operatorname{Attention}$ in Transformers has a significantly smaller impact than removing $\operatorname{Mlp}$. We progressively removed attention or MLP layers, starting from the last layer while retaining the first layer. Fig.~\ref{fig:pca_skip_analysis} (Left) demonstrates that for ResFormer, the impact of removing $\operatorname{Attention}$ is more comparable to that of removing $\operatorname{Mlp}$, in contrast to vanilla Transformers. This indicates that the $\operatorname{Attention}$ in ResFormer, with value residual, contribute more significantly to each layer's hidden states than in vanilla Transformers.

Furthermore, we performed PCA dimensionality reduction on the hidden states of each layer in both ResFormer and vanilla Transformer models. We determined the minimum number of principal components required to explain $99\%$ of the variance. Fig.~\ref{fig:pca_skip_analysis} reveals that ResFormer, starting from the second layer where value residual connections are introduced, consistently produces hidden states with a higher minimum number of principal components compared to vanilla Transformers. This suggests that ResFormer generates hidden states with higher information density.

\subsection{SVFormer vs. GQA,CLA}
\begin{table}[!h]
    \centering
    \begin{subtable}{0.46\linewidth}
        \centering
        \resizebox{\linewidth}{!}{\begin{tabular}{ccc}
            \toprule
Sequence & \multirow{2}{*}{Model} & Valid \\
Length &  &  Loss\\
            \midrule
            \multirow{4}{*}{2,048} & - & 2.739 \\
             & CLA2 &  2.776 \\
             & GQA2 &  2.748 \\
             & SVFormer & 2.774  \\
            \multirow{4}{*}{64,000} & - & 2.753 \\
             & CLA2 &  2.793 \\
             & GQA2 &  2.773 \\
             & SVFormer & 2.7485  \\
            \bottomrule
        \end{tabular}}
    \end{subtable}
    \hspace{0.02\linewidth}
    \begin{subtable}{0.46\linewidth}
        \centering
        \resizebox{\linewidth}{!}{\begin{tabular}{ccc}
            \toprule
Sequence & \multirow{2}{*}{Model} & Valid \\
Length &  &  Loss\\
            \midrule
            \multirow{6}{*}{64,000} & - & 2.753 \\
             & GQA8 &  2.807 \\
             & \makecell{CLA2\\\quad\quad+GQA4} &  2.815 \\
             & \makecell{SVFormer\\\quad\quad+GQA4} & 2.741  \\
            \bottomrule
        \end{tabular}}
    \end{subtable}
    \caption{Comparison of valid loss under varying degrees of $KV$ cache reduction. CLA2 denotes parameter sharing every two layers, while GQA2 indicates halving the key-value heads. Left: Model with nearly ${1}/{2}$ $KV$ cache. Right: Model with nearly ${1}/{8}$ $KV$ cache.}
    \label{tab:SVattention_kvcache}
\end{table}

\begin{figure}[!h]
      \centering
      \begin{subfigure}{0.47\linewidth}
        \centering
        \includegraphics[width=\linewidth]{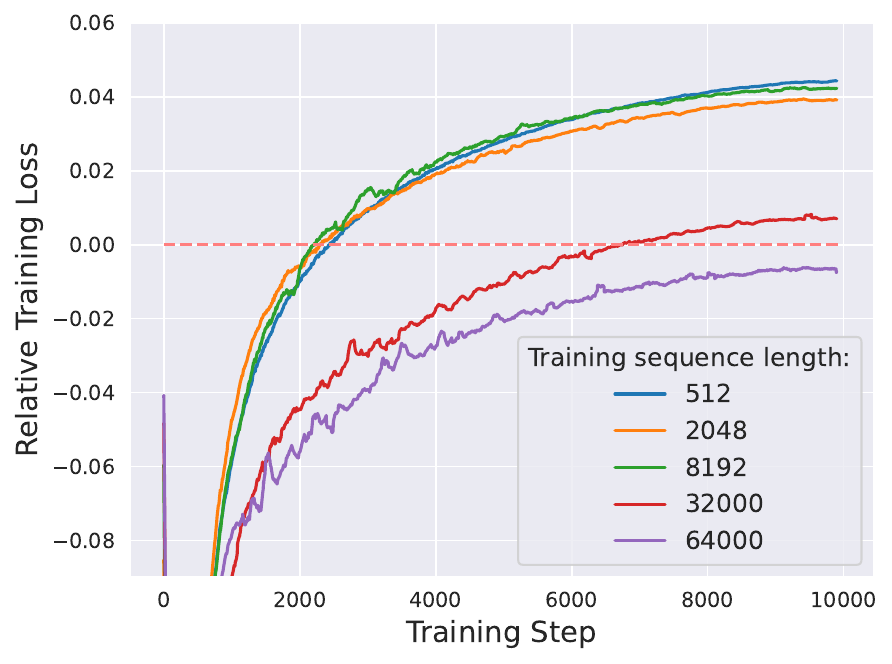}
      \end{subfigure}
      \begin{subfigure}{0.47\linewidth}
        \centering
        \includegraphics[width=\linewidth]{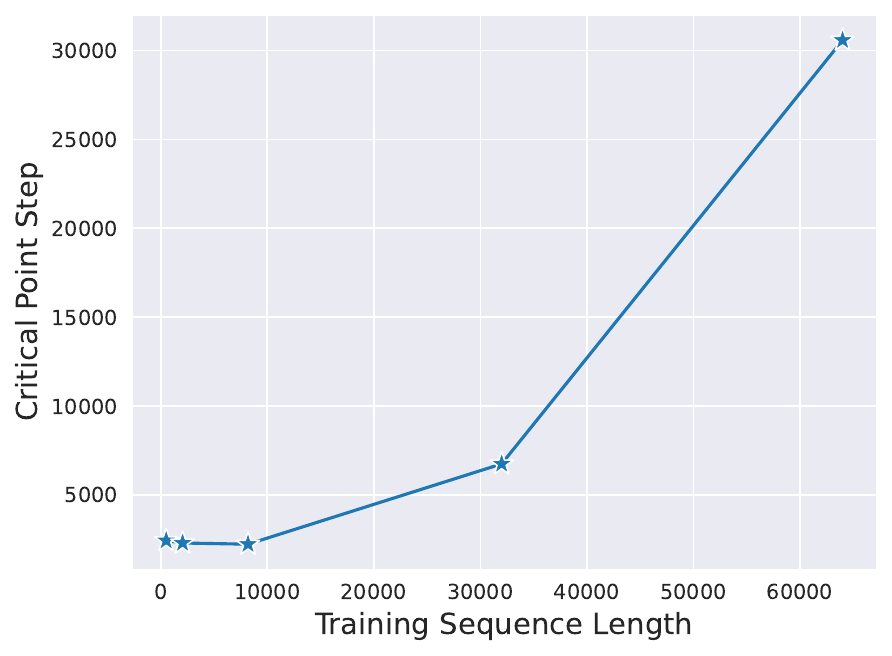}
      \end{subfigure}
      \caption{Left: Relative training loss for SVFormer $vs.$ vanilla Transformer under different sequence lengths with a fixed batch size of 2M tokens. Right: Analysis of critical point, and we predict it for length 64,000 using linear regression with the last 1,000 data points.}
      \label{fig:SVattention_seqlength}
\end{figure}

\begin{figure}[htbp]
\small
\centering
\begin{minipage}{0.46\linewidth}
    \centering
    \resizebox{\linewidth}{!}{\begin{tabular}{cccc}
        \toprule
        \makecell{Model\\Type} & \makecell{Learning\\Rate} & \makecell{Warnup\\Steps}  & \makecell{Valid\\Loss} \\
        \midrule
        \multirow{4}{*}{Llama} & 1e-4 & 120 & -0.033 \\
         & 3e-4 & 120 & +0.021 \\
         & 6e-4 & 120 & +0.035 \\
         & 6e-4 & 1,200 & +0.036 \\
        GPT2 & 6e-4 & 120 & +0.029 \\
        \bottomrule
    \end{tabular}}
    \captionof{table}{Relative validation loss of SVFormer compared to vanilla Transformer under different hyper-parameter settings.}
    \label{tab:svformer_other_factors}
\end{minipage}%
\hspace{0.03\linewidth}
\begin{minipage}{0.46\linewidth}
    \centering
    \includegraphics[width=\linewidth]{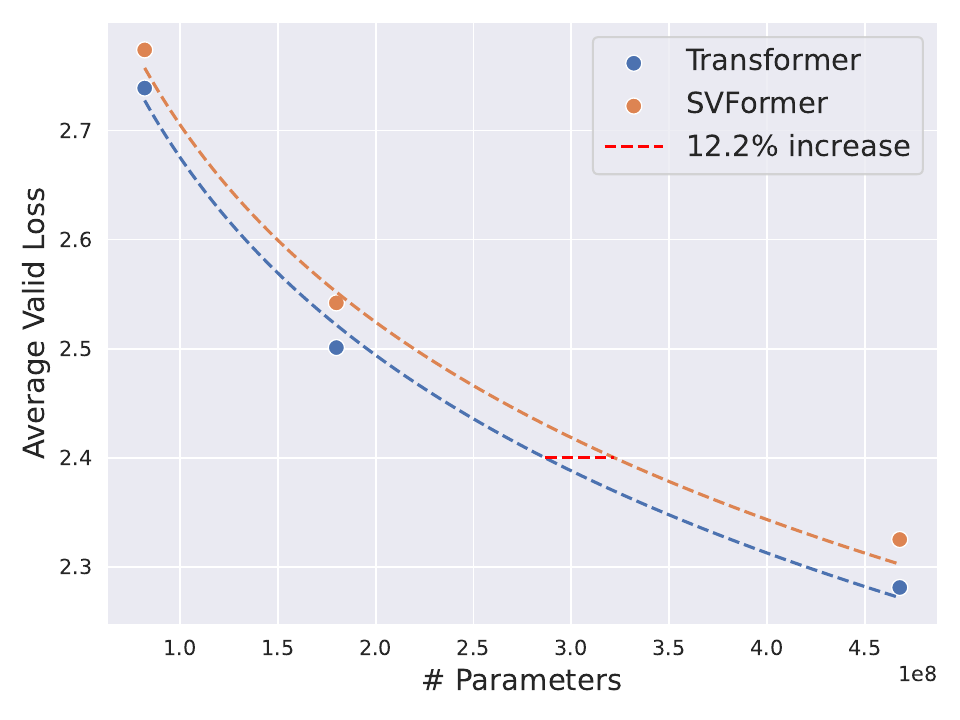}  
    \caption{Validation loss for SVFormer as model size scales from 82M to 468M.}
    \label{fig:svformer_modelsize}
\end{minipage}
\end{figure}

In the Table~\ref{tab:SVattention_kvcache}, at a training sequence length of 64,000, SVFormer demonstrates lower final loss compared to existing $KV$-efficient methods such as CLA and GQA. Moreover, it can be used concurrently with GQA to enhance $KV$ efficiency further. However, we observed that with a training sequence length of 2,048, SVFormer underperforms compared to GQA. The results indicate that sequence length significantly affects SVFormer's performance. Thus, we conducted more comprehensive experiments on sequence length.

\paragraph{Effects of sequence length} Results in Fig.~\ref{fig:SVattention_seqlength} (Left) demonstrate that SVFormer will always be gradually surpassed by vanilla attention during training while its training speed is faster than vanilla Transformer at the early stage. However, as the training sequence length increases, the SVFormer model performs better. In this way, we focus on the critical point, defined as the number of training steps exceeded. Fig.~\ref{fig:SVattention_seqlength} (Right) illustrates that the relationship between the critical point and sequence length exhibits an exponential trend. We argue that it's due to the challenge deep models face in fully optimizing the increasingly larger first-layer value matrix as the sequence length grows.

\paragraph{Other factors} Table~\ref{tab:svformer_other_factors} show SVFormer benefits more from smaller learning rates than from warmup. This aligns with performance correlating to total summed learning rate \citep{kaplan2020scaling}. Larger models, requiring smaller learning rates, suit SVFormer better. Results also indicates the SVFormer-Transformer difference is not architecture-sensitive. Compared with Transformer, SVFormer requires a 12.2\% increase in parameters to achieve the same loss while reducing the $KV$-cache by nearly half in Fig.~\ref{fig:svformer_modelsize}.

\section{Conclusion}
In this paper, we demonstrate the inadequacy of existing hidden residual connections in propagating information from the initial token-level to deeper layers. To address this limitation, we propose ResFormer, which incorporates a residual connection between the value vectors of the current layer and those of the first layer prior to the attention operation. Furthermore, we introduce SVFormer, an extension of ResFormer, which achieves a nearly 50\% reduction in the $KV$ cache. We conducted extensive experiments on language modeling tasks to evaluate the efficacy of these two Transformer variants across diverse scenarios.

\section*{Limitations}
The proposed learnable ResFormer, still falls short of identifying the optimal $\bm\lambda$ setting through current training, instead converging on a relative optimum. This limitation suggests that further refinement of initialization strategies and learning algorithms may be necessary. Due to computational constraints, we were unable to conduct experimental validation on larger-scale models at this time.

\section*{Ethics Statement}
On the one hand, the data employed in this paper is sourced from publicly available datasets provided by the company, which have undergone a certain level of filtering. On the other hand, the models trained in our study are solely utilized for experimental analysis and will not be publicly deployed.

\section*{Acknowledgments}
This work was supported by the Scientific Research Project of Westlake University (Grant No. WU2024B003).

\bibliography{custom}

\appendix

\section{Appendix}
\label{sec:appendix}
\subsection{Attention pattern analysis}
\begin{figure}[!h]
    \begin{minipage}{0.48\linewidth}
        \centering
        \includegraphics[width=\linewidth]{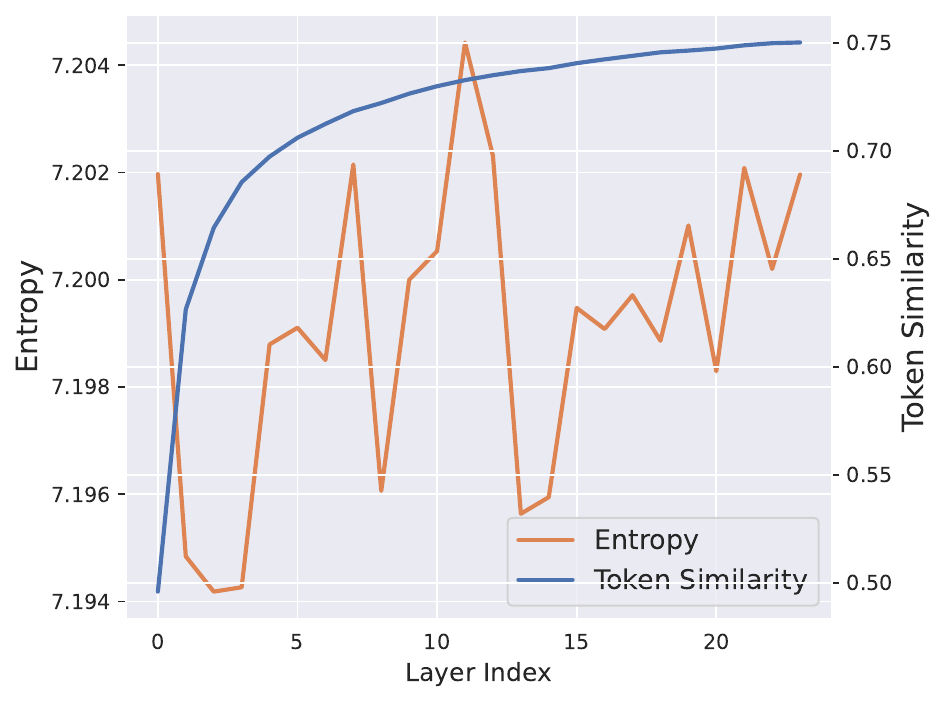}
    \end{minipage}
    \hfill
    \begin{minipage}{0.48\linewidth}
        \centering
        \includegraphics[width=\linewidth]{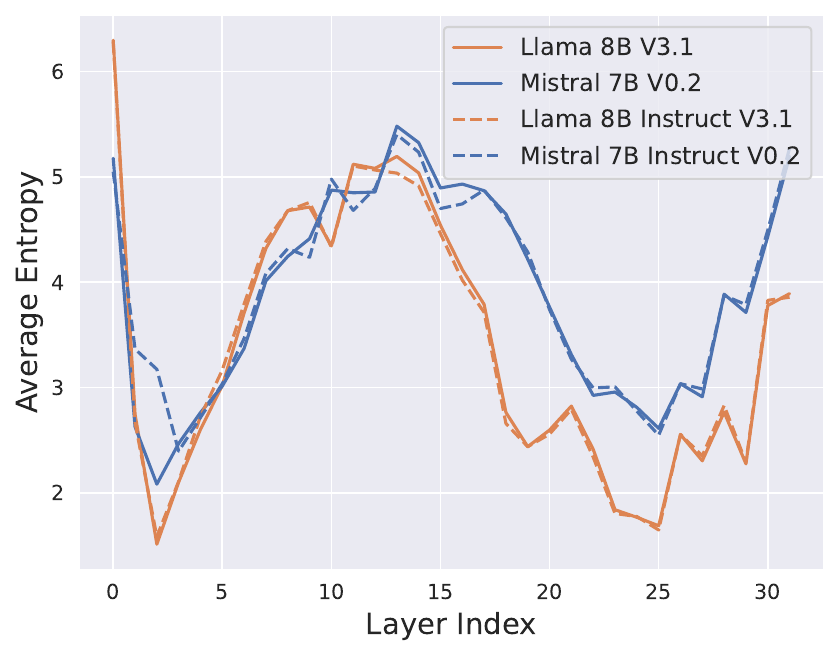}
    \end{minipage}
    \vspace{-2pt}
    \caption{(Left) Average entropy of token importance and the average hidden-state similarity for a randomly initialized 468M model. (Right) Average entropy of token importance across layers in Llama (8B) \citep{dubey2024llama} and Mistral (7B) \citep{jiang2023mistral}.}
    \label{fig:attention_concentrate}
    \vspace{-16pt}
\end{figure}
\begin{figure}[!h]
  \centering
\begin{subfigure}{0.48\linewidth}
\centering
\includegraphics[trim=0 0 0 0,clip,width=\linewidth]
{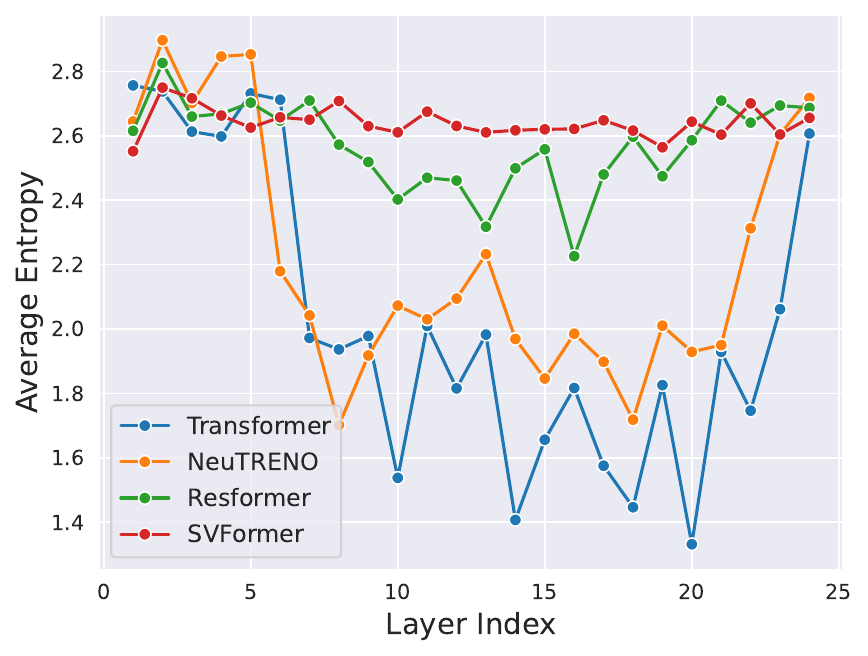}
\caption{Attention Entropy.}
\label{fig:attention_entropy}
\end{subfigure}
\hfill
\begin{subfigure}{0.48\linewidth}
\centering
\includegraphics[trim=0 0 0 0,clip,width=\linewidth]
{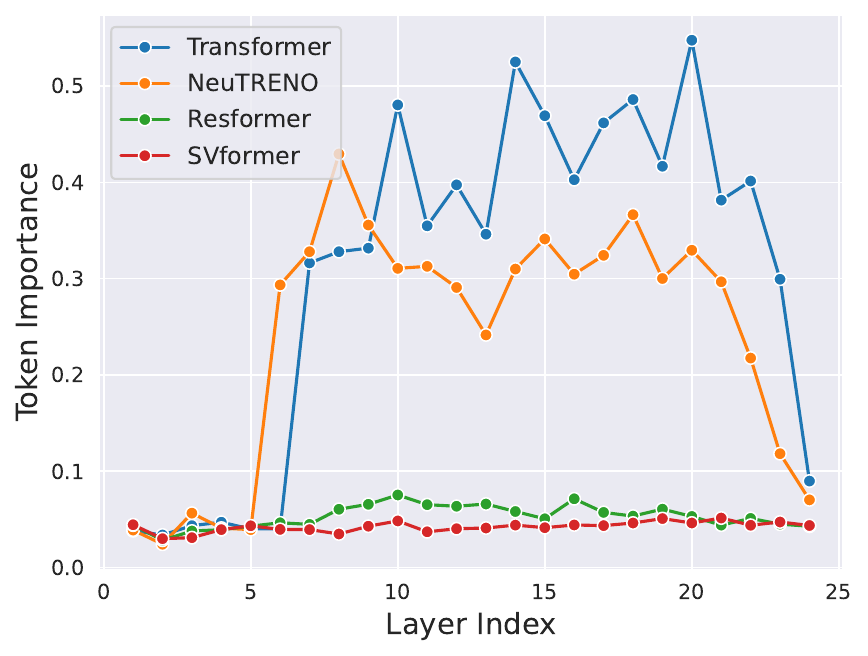}
\caption{Token importance.}
\end{subfigure}
\begin{subfigure}{0.48\linewidth}
\centering
\includegraphics[trim=0 0 0 0,clip,width=\linewidth]
{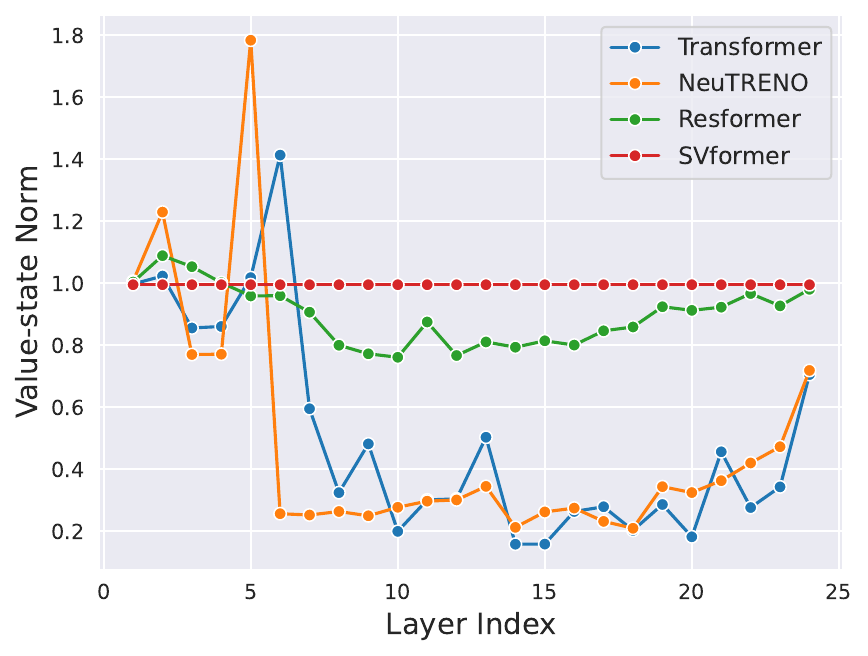}
\caption{Norms of value states.}
\end{subfigure}
\hfill
\begin{subfigure}{0.48\linewidth}
\centering
\includegraphics[trim=0 0 0 0,clip,width=\linewidth]
{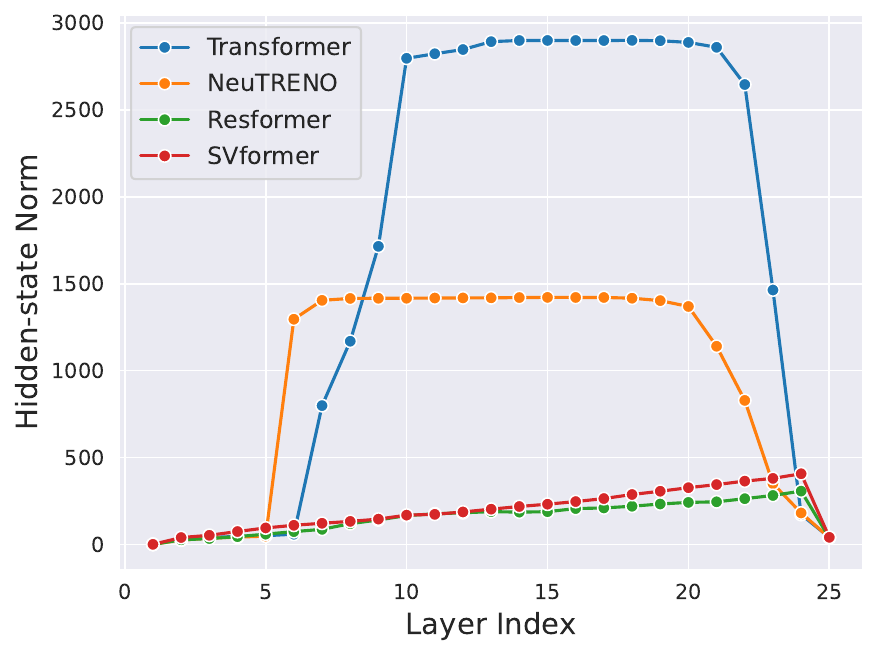}
\caption{Norms of hidden states.}
\end{subfigure}
\vspace{-2mm}
\caption{The token importance~\citep{xiao2023efficient}, value-state norms~\citep{guo2024attention}, and hidden-state norms~\citep{sun2024massive} of the first token across layers of 468M models. ``Attention Entropy" refers to the entropy of token importance across each sequence. }
\label{fig:sink-value-hidden}
\end{figure}
\paragraph{Attention concentration}
Given the attention matrix $\mathbf{A}  \in \mathbb{R}^{l \times l}$ at one layer, we use entropy $\bm{e}$ to represent its concentration effect. To obtain entropy $\bm{E}$, calculate the importance vector $\mathbf{a} = \frac{1}{l}\sum_{j = 1}^{l}A_{ij}$ firstly where $\mathbf{A}$ is a lower triangular matrix. The entropy can be formulated as follows: $\bm{e} = - \sum_{i=1}^{l} \bm{a}_i^{\prime} \log{\bm{a}^{\prime}_i}$, where $a_i^{{\prime}} = {a_i}/\Big({\sum_{i=1}^{l}{a_i}}\Big)$ for $i = 1, 2, \dots, l$ and the higher the entropy $\bm{e}$, the greater the degree of clustering in $\bm{a}$, i.e., attention matrix $\mathbf{A}$ is more likely to focus on specific tokens.

The phenomenon of attention concentration is inherent to model architecture and emerges during training. Fig.~\ref{fig:attention_concentrate} shows that randomly initialized models exhibit over-smoothing but not attention concentration and popular trained models exhibit obvious attention concentration problem. Trained ViT models often focus on low-informative background areas \citep{darcet2023vision}, while language models concentrate on low-semantic tokens \citep{sun2024massive}, particularly the start token (attention sink \citep{xiao2023efficient}). While previous studies analyzed single-layer attention patterns, our research reveals a ``concentration - dispersion - concentration" pattern in deep models, as shown in Fig.~\ref{fig:attention_concentrate} (Right), suggesting potential loss of information during concentrated phases.

\paragraph{ResFormer alleviates attention concentration}
\begin{figure*}
    \centering
    \begin{minipage}{0.13\textwidth}
        \footnotesize
        \centering
        Transformer 
    \end{minipage}%
    \begin{minipage}{0.29\textwidth}
        \centering
        \includegraphics[width=\linewidth]{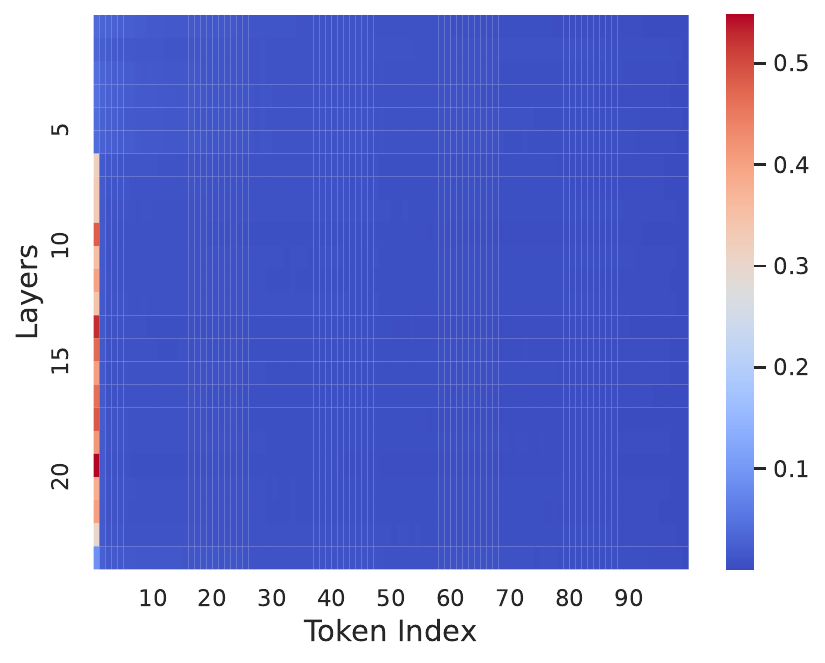}
    \end{minipage}%
    \begin{minipage}{0.29\textwidth}
        \centering
        \includegraphics[width=\linewidth]{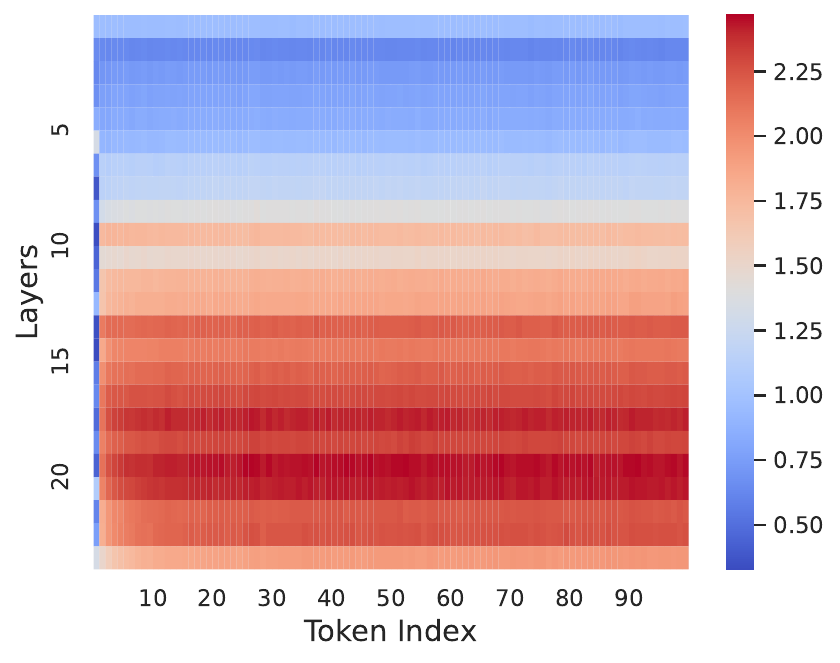}
    \end{minipage}%
    \begin{minipage}{0.29\textwidth}
        \centering
        \includegraphics[width=\linewidth]{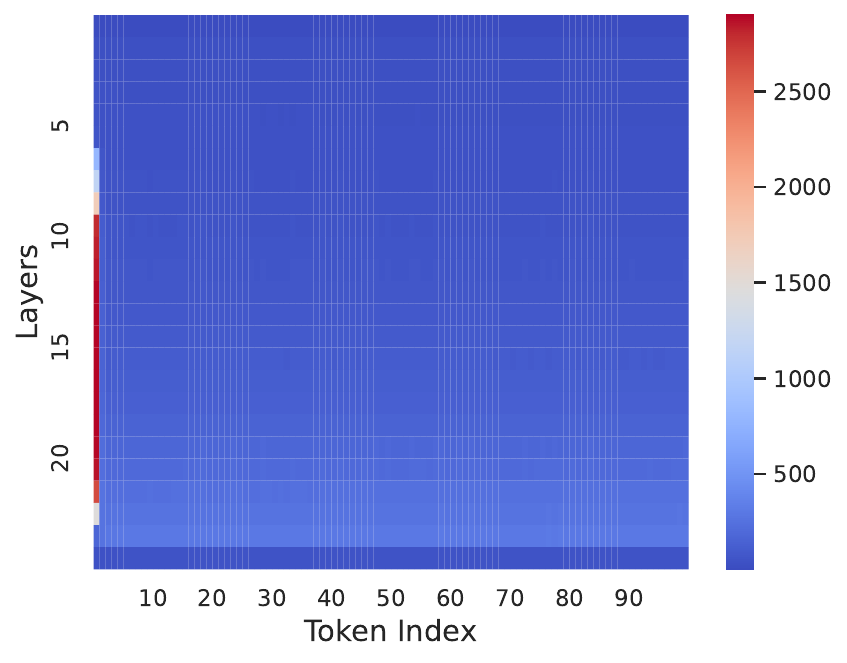}
    \end{minipage}

    \begin{minipage}{0.13\textwidth}
        \footnotesize
        \centering
        NeuTRENO
    \end{minipage}%
    \begin{minipage}{0.29\textwidth}
        \centering
        \includegraphics[width=\linewidth]{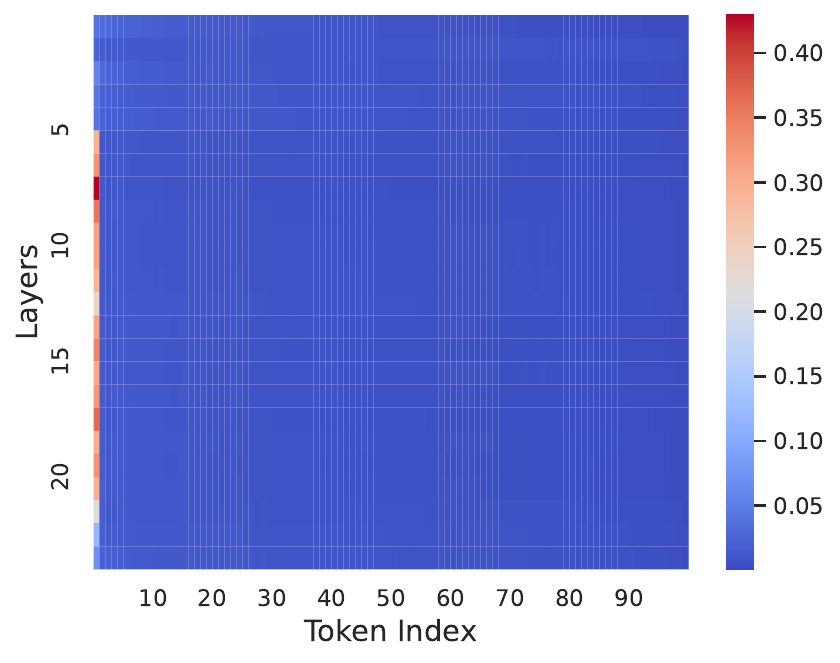}
    \end{minipage}%
    \begin{minipage}{0.29\textwidth}
        \centering
        \includegraphics[width=\linewidth]{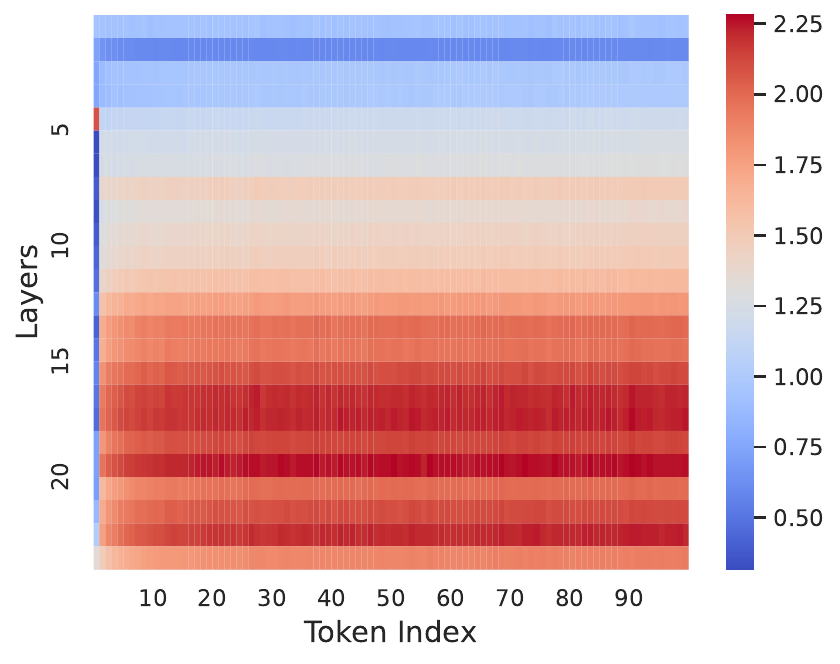}
    \end{minipage}%
    \begin{minipage}{0.29\textwidth}
        \centering
        \includegraphics[width=\linewidth]{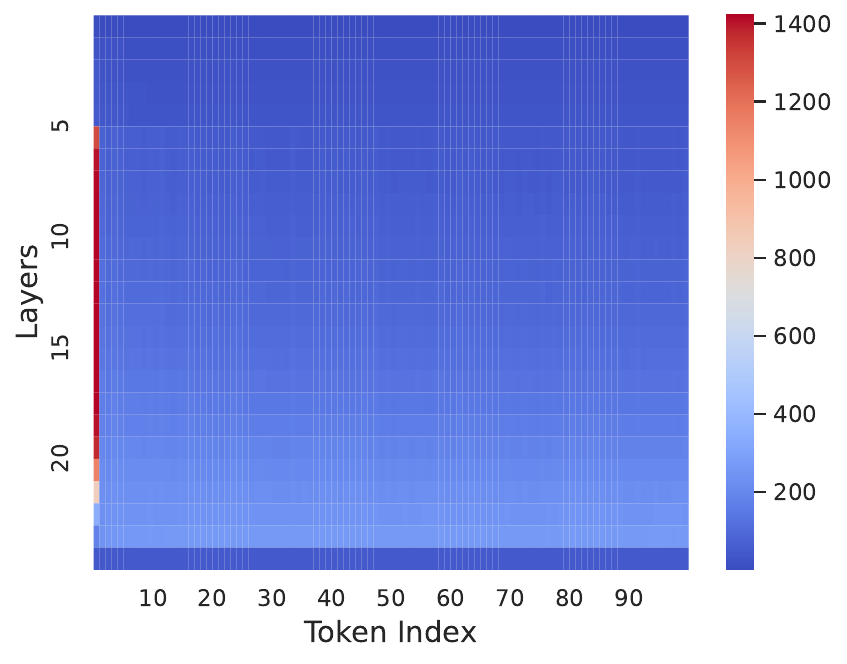}
    \end{minipage}

    \begin{minipage}{0.13\textwidth}
        \footnotesize
        \centering
        ResFormer
    \end{minipage}%
    \begin{minipage}{0.29\textwidth}
        \centering
        \includegraphics[width=\linewidth]{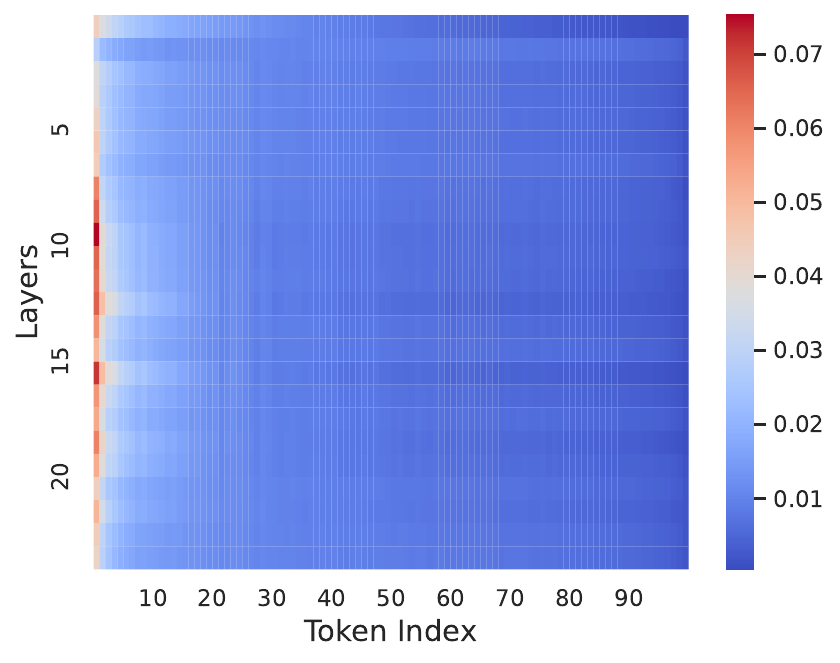}
    \end{minipage}%
    \begin{minipage}{0.29\textwidth}
        \centering
        \includegraphics[width=\linewidth]{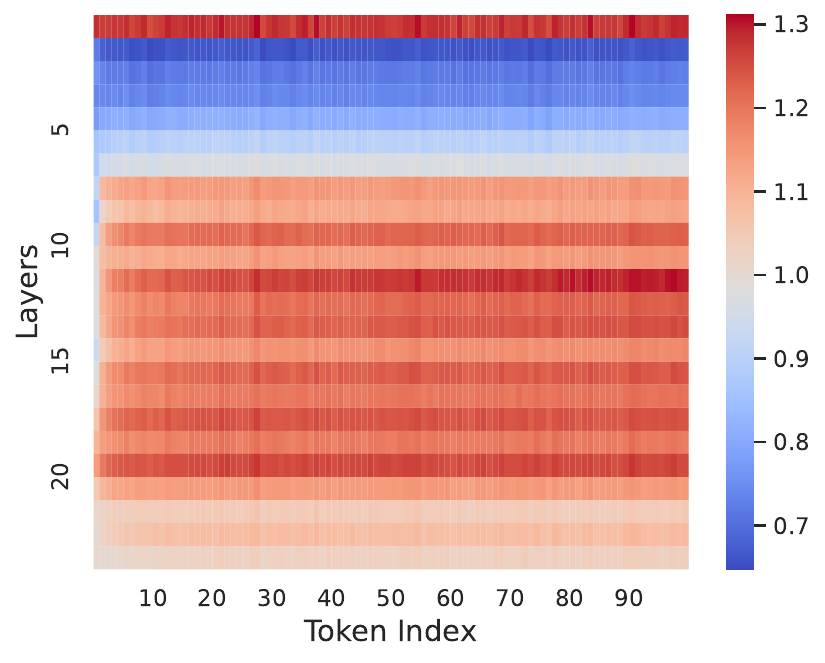}
    \end{minipage}%
    \begin{minipage}{0.29\textwidth}
        \centering
        \includegraphics[width=\linewidth]{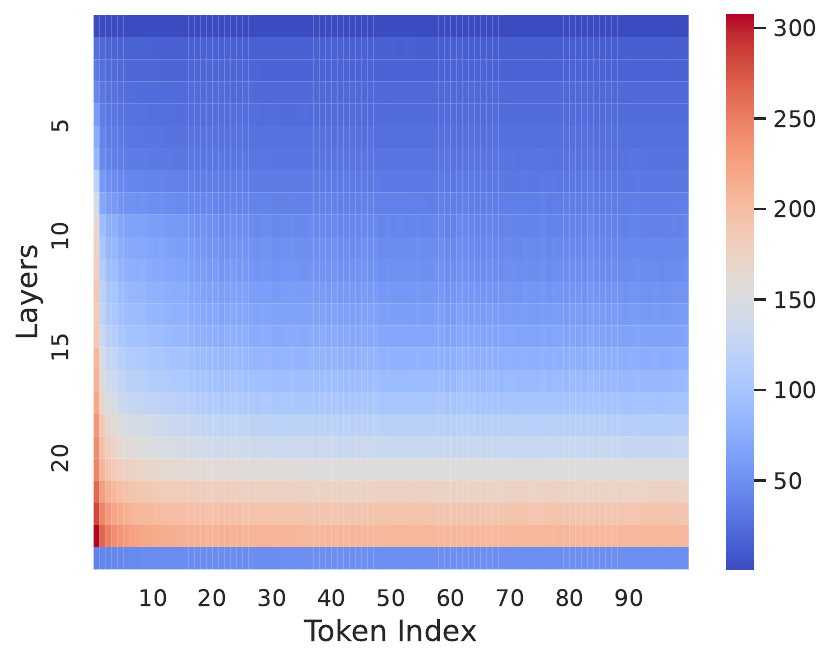}
    \end{minipage}

    \begin{minipage}{0.13\textwidth}
        \footnotesize
        \centering
        SVFormer
    \end{minipage}%
    \begin{minipage}{0.29\textwidth}
        \centering
        \includegraphics[width=\linewidth]{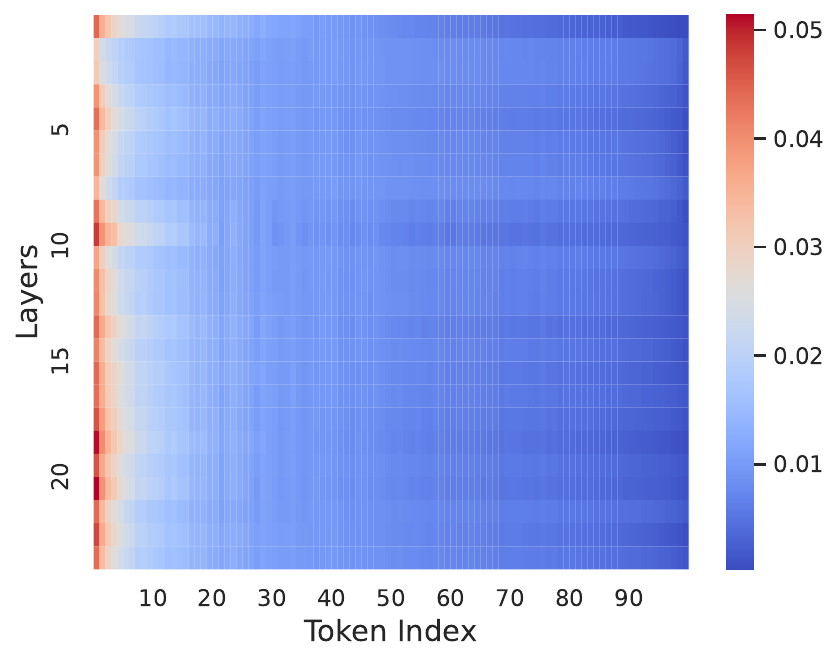}
        \subcaption{Token importance.}
    \end{minipage}%
    \begin{minipage}{0.29\textwidth}
        \centering
        \includegraphics[width=\linewidth]{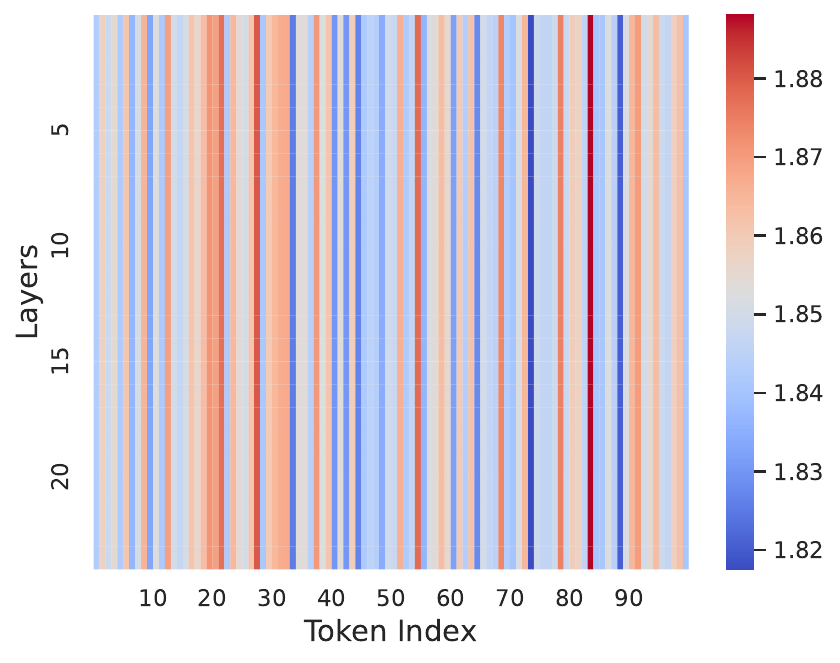}
        \subcaption{Value-state norms.}
    \end{minipage}%
    \begin{minipage}{0.29\textwidth}
        \centering
        \includegraphics[width=\linewidth]{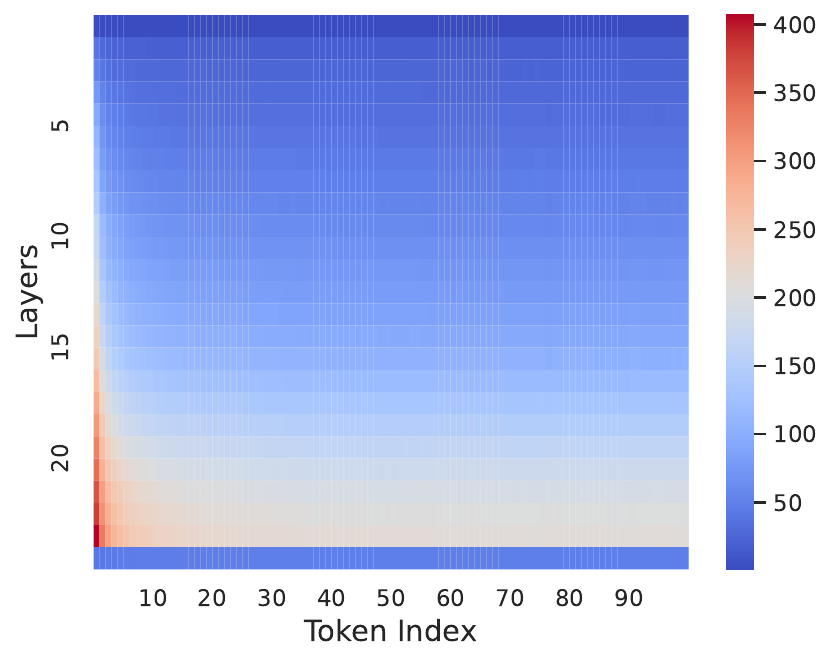}
        \subcaption{Hidden-state drains.}
    \end{minipage}
\caption{Visualization of token importance, value state norms, and hidden state norms across different token positions and layers in 468M models.}
\label{fig:heatmap}
\end{figure*}
\begin{figure*}[!h]
  \centering
\begin{subfigure}{0.24\linewidth}
\centering
\includegraphics[trim=0 0 0 0,clip,width=\linewidth]{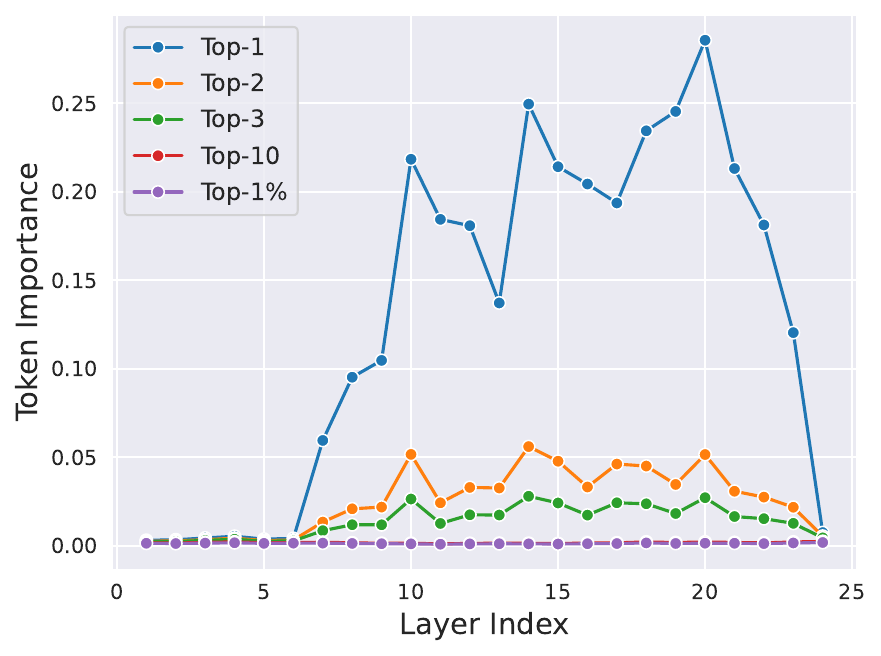}
\caption{Transformer.}
\end{subfigure}
\begin{subfigure}{0.24\linewidth}
\centering
\includegraphics[trim=0 0 0 0,clip,width=\linewidth]{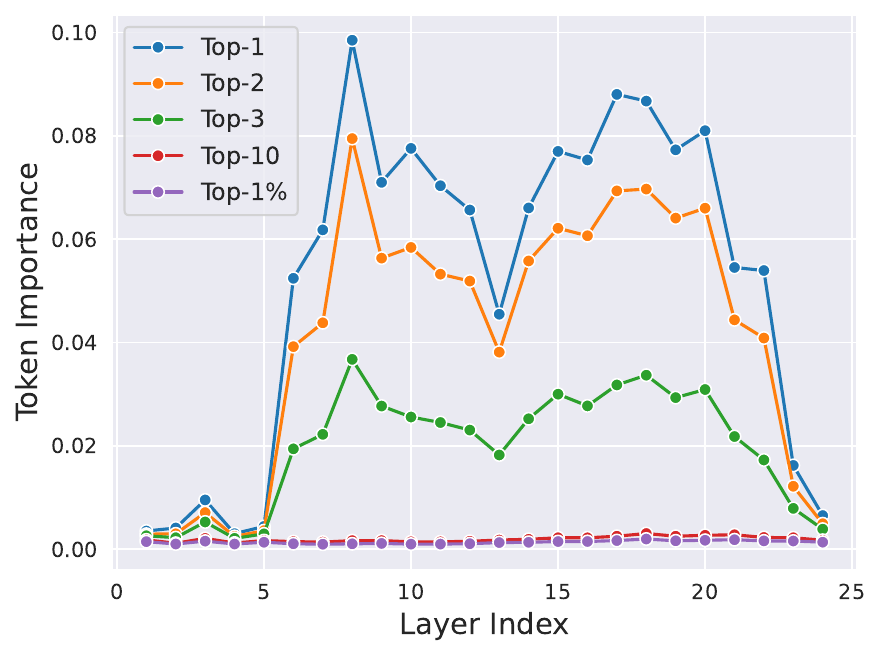}
\caption{NeuTRENO.}
\end{subfigure}
\begin{subfigure}{0.24\linewidth}
\centering
\includegraphics[trim=0 0 0 0,clip,width=\linewidth]{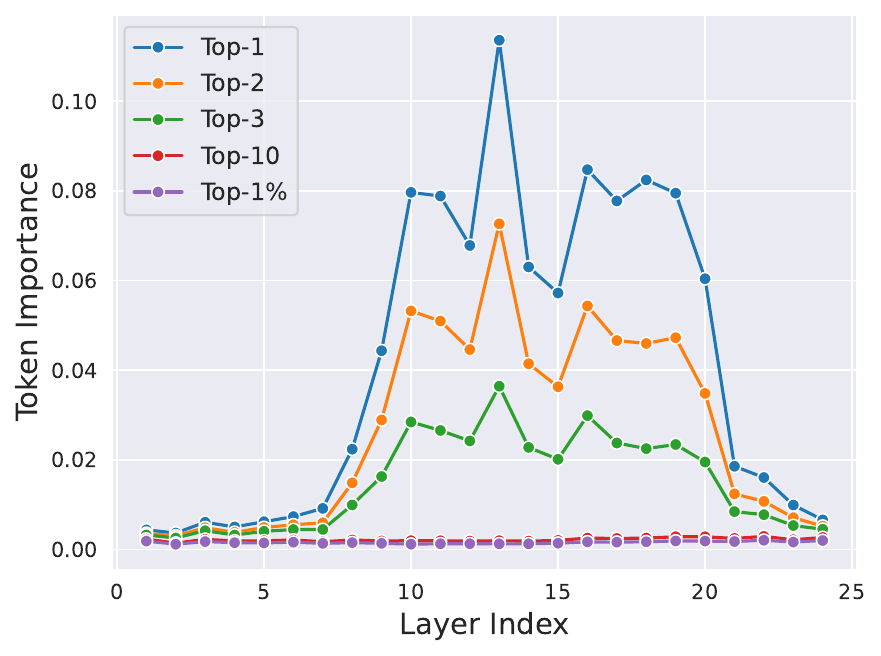}
\caption{ResFormer.}
\end{subfigure}
\begin{subfigure}{0.24\linewidth}
\centering
\includegraphics[trim=0 0 0 0,clip,width=\linewidth]{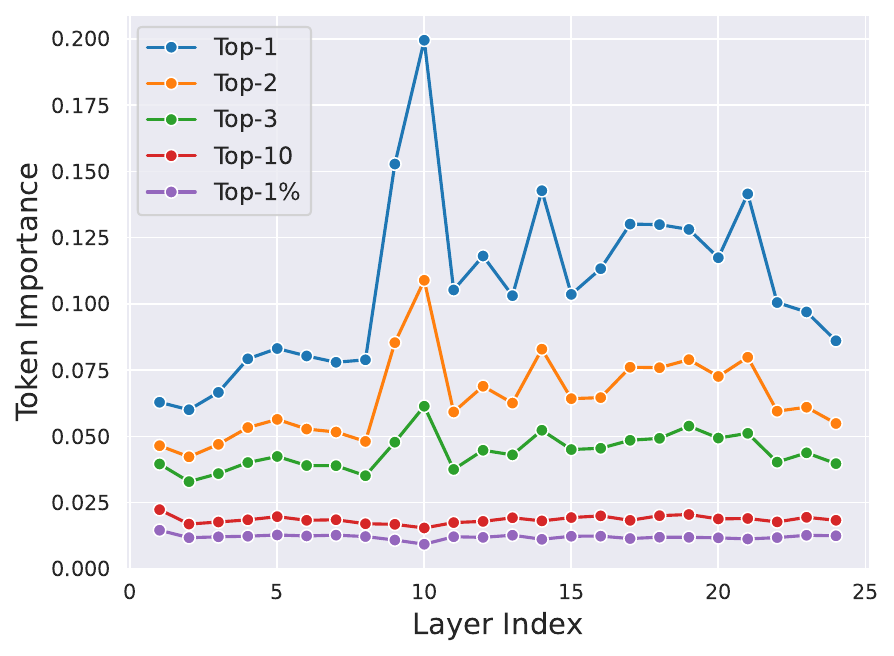}
\caption{SVFormer.}
\end{subfigure}
\caption{The distribution of token importance for different models at different layers.}
\label{fig:attention_distribution}
\vspace{-6pt}
\end{figure*}
Fig.~\ref{fig:attention_entropy} illustrates that the clustering effect of attention increases significantly with the number of layers for the vanilla Transformer, whereas the clustering effect is relatively less pronounced for the ResFormer. We further visualize the attention weights, value-state norms $\| \bm{v} \|_2$, and hidden-state norms $\| \bm{h} \|_2$ of tokens at different layers and positions. Given that attention clustering often occurs on the first token, we primarily show its results in Fig.~\ref{fig:sink-value-hidden}. The results indicate that using ResFormer significantly mitigates attention sinks \citep{xiao2023efficient}, value-state drains \citep{guo2024attention} and residual-state peaks \citep{sun2024massive}. \cite{guo2024active} attributes these phenomena to the mutual reinforcement mechanism of model between value-state drains and attention sinks. We suggest that the value shortcut disrupts this mechanism by alleviating value-state drains due to the absence of value-state drains in the first layer. Specifically, for tokens lacking semantic information like start tokens, a large value state magnitude can adversely affect the prediction of subsequent tokens if they are overly attended to since $\mathbf{U}_n = \mathbf{A}_n\mathbf{V}_n$ in Eqn.\ref{eqn:attention}. However, when there is no value-state drains, models will reduce attention clustering to these tokens to minimize loss.

Fig.~\ref{fig:heatmap} (First column) demonstrates that the start token easily attracts massive attention despite lacking semantic information for Transformer and NeuTRENO. And Fig.\ref{fig:attention_distribution} further illustrates the distribution of token importance, where TOP-$i$ represents the $i$-th largest token importance within a sequence. Compared to Transformer and NeuTRENO, ResFormer and SVFormer exhibit a more uniform distribution of token importance.

\paragraph{Any negative effect?}
\begin{table}[!h]
\centering
\small
\renewcommand{\arraystretch}{1.25}
\begin{tabular}{ccc}
\toprule
$\lambda$ value & Layers & Valid loss \\
\midrule
- & - & 2.739 \\
\multirow{3}{*}{1} & All layers & 2.7037 \\
& 2,3,4 & 2.724 \\
& 5,6,7 & 2.698 \\
\hline
\multirow{3}{*}{2} & All layers & 2.7 \\
& 2,3,4 & 2.728 \\
& 5,6,7 & 2.688 \\
\hline
\multirow{3}{*}{5} & All layers & 2.704 \\
& 2,3,4 & 2.732 \\
& 5,6,7 & 2.682 \\
\bottomrule
\end{tabular}
\caption{Performance of Sparse ResFormer with different $\lambda$ values and different layer configurations.}
\label{tab:resformer_sparsity_v1}
\end{table}
\begin{table}[!h]
\centering
\small
\renewcommand{\arraystretch}{1.25}
\begin{tabular}{ccc}
\toprule
ResFormer Layers & Norm re-scale & Valid loss \\
\midrule
\multirow{2}{*}{All Layers} & No & 2.712 \\
 & Yes & 2.701 \\
\hline
\multirow{2}{*}{2,3,4} & No & 2.734 \\
 & Yes & 2.727 \\
\hline
\multirow{2}{*}{6,7,8} & No & 2.702 \\
 & Yes & 2.701 \\
\bottomrule
\end{tabular}
\caption{Ablation study of post-value residual re-scaling.}
\label{tab:resformer_sparsity_v2}
\end{table}
Mitigating attention concentration may enhance interpretability but potentially affect transformer sparsity. As attention sinks typically emerge early, applying value residual to later layers should have less impact intuitively. We compared two sparse ResFormer variants on an 8-layer model, applying value residual to layers 2-4 versus 6-8. The results in Table~\ref{tab:resformer_sparsity_v1} demonstrate that while incorporating value residual generally improves performance compared to vanilla Transformers, increasing $\lambda$ (the proportion of $\mathbf{V}_1$) in the value residual led to decreased performance for shallower networks.  Conversely, deeper networks showed improved results with higher $\lambda$ values. Notably, the Learnable-ResFormer learned to apply value residual primarily to later layers, minimizing the impact on network sparsity. Moreover, we implemented post-value residual re-scaling ($\mathbf{V}_{n}^{\prime} =  \frac{\|\mathbf{V}_{n}\|}{\|0.5\mathbf{V}_{1}+0.5\mathbf{V}_{n}\|}({0.5}\mathbf{V}_{1}+{0.5}\mathbf{V}_{n})$) to mitigate its impact on later layers. This approach benefited shallow-sparse ResFormers but had minimal effect on deep-sparse variants. This further suggests that the value residual patterns learned by the Learnable ResFormer do not introduce significant negative effects in this context.


\subsection{Pre-train Dataset}

\begin{table*}[htbp]
\centering
\resizebox{0.4\linewidth}{!}{%
\begin{tabular}{@{}ccc@{}}
\toprule
Data source & proportions & Tokens \\ \midrule
Commoncrawl & 50\% & 10 B \\
C4 & 20\% & 4 B \\
GitHub & 10\% & 2 B \\
Books & 5\% & 1 B \\
ArXiv & 5\% & 1 B \\
Wikpedia & 5\% & 1 B \\
StackExchange & 5\% & 1 B \\ \bottomrule
\end{tabular}%
}
\caption{The details of pre-train dataset.}
\label{tab:slimpajama-data-proportion}
\end{table*}

Based on the equation $D \geq 5000 \cdot N^{0.74}$ \citep{kaplan2020scaling} where $D$ is data size and $N$ is the number of non-embedding parameters, we need to collect at least 17.5B for model has N = 700M non-embedding parameters (corresponding to complete 1B model with 2,048 hidden size, 50,277 vocab size and 2,048 sequence length) to avoid over-fitting. Besides, \cite{xie2024doremi} indicates that the mixture proportions of pre-training data domains significantly affects the training results. In this way, we sampled 20B tokens data from original 627B data based on the original data proportions shown in the Table~\ref{tab:slimpajama-data-proportion}.

\subsection{Training Details}
\label{sec:training_details_appendix}
\begin{table*}[htbp]
\centering
\renewcommand{\arraystretch}{1.25}
\resizebox{0.7\linewidth}{!}{%
\begin{tabular}{@{}lccccc@{}}
\toprule
Max Sequence Length         & 512 & 2,048 & 8,192 & 32,000 & 64,000 \\ \midrule
Total Batch Size  &  4,096   &  1,024    &   256   &    64   &   32    \\ 
Per-GPU Batch Size          & 128 &   32   &   8   &   2    &  1     \\
Gradient Accumulation Step  & \multicolumn{5}{c}{32}       \\ 
GPUs              & \multicolumn{5}{c}{8}       \\ 
\bottomrule
\end{tabular}
}
\caption{{Training details for training dataset with different sequence length.}}
\label{tab:training_detail_data}
\vspace{-5pt}
\end{table*}
Section~\ref{sec:training_details} introduces the main experimental hyperparameters used in the paper. This section further details the training parameters for various model sizes and training sequence lengths. Table~\ref{tab:training_detail_model} demonstrates the differences among models of various sizes. The configurations for the number of layers, attention heads, hidden dimensions, and FFN dimensions are based on \cite{biderman2023pythia}. Moreover, as reported in Table~\ref{tab:training_detail_data}, the batch size that a single GPU can accommodate varies depending on the length of the training sequences. Note that the total number of tokens in each batch is consistently 2 million.

\begin{table*}[htbp]
\centering
\renewcommand{\arraystretch}{1.25}
\resizebox{0.8\linewidth}{!}{%
\begin{tabular}{@{}lcccc@{}}
\toprule
Model Size                                & 2M & 82M & 180M & 468M \\ \midrule
Layers                                    &  4  &  8   &  12    &  24    \\
Attention Heads                           &  2  &  8   &  12    &  16    \\
Hidden Dimension                          &  16  & 512    &  768    &  1,024    \\
FFN Dimension                              &  56  & 1,792    & 2,688    &  3,584 \\
Tie Word Embedding                        & \multicolumn{4}{c}{False}    \\
(Peak Learning Rate, Final Learning Rate) & \multicolumn{4}{c}{$(6e-4, 6e-5)$}   \\
Learning Rate Schedule                    & \multicolumn{4}{c}{Cosine Decay}  \\
Vocabulary Size                           & \multicolumn{4}{c}{50,277}      \\
Activation Function                       & \multicolumn{4}{c}{SwiGLU}      \\
Position Embedding                  & \multicolumn{4}{c}{RoPE ($\theta$ = 10,000)} \\
Batch Size                               & \multicolumn{4}{c}{2M tokens}      \\
Data Size                               & \multicolumn{4}{c}{20B tokens}      \\
(Warmup Steps, Training Steps)            & \multicolumn{4}{c}{(120, 10,000)}  \\
Adam $\beta$                                & \multicolumn{4}{c}{(0.9, 0.95)} \\
Dropout                                   & \multicolumn{4}{c}{0.0}      \\
Weight Decay                               & \multicolumn{4}{c}{0.1}      \\ \bottomrule
\end{tabular}
}
\caption{Training details for models with different size.}
\label{tab:training_detail_model}
\end{table*}

\end{document}